\documentclass[sigconf]{acmart}

\usepackage{balance}
\usepackage{kotex}
\usepackage{multirow}
\usepackage{makecell}
\usepackage{xcolor}
\usepackage{arydshln}
\usepackage{wrapfig}
\usepackage{pifont}

\usepackage{amsmath}
\usepackage{amsthm}
\usepackage{booktabs}
\usepackage{algorithm}

\usepackage{algpseudocode}%

\AtBeginDocument{%
  }

\copyrightyear{2024}
\acmYear{2024}
\setcopyright{rightsretained}
\acmConference[CIKM '24]{Proceedings of the 33rd ACM International Conference on Information and Knowledge Management}{October 21--25, 2024}{Boise, ID, USA}
\acmBooktitle{Proceedings of the 33rd ACM International Conference on Information and Knowledge Management (CIKM '24), October 21--25, 2024, Boise, ID, USA}
\acmDOI{10.1145/3627673.3679526}
\acmISBN{979-8-4007-0436-9/24/10}

\makeatletter
\gdef\@copyrightpermission{
  \begin{minipage}{0.3\columnwidth}
   \href{https://creativecommons.org/licenses/by/4.0/}{\includegraphics[width=0.90\textwidth]{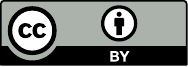}}
  \end{minipage}\hfill
  \begin{minipage}{0.7\columnwidth}
   \href{https://creativecommons.org/licenses/by/4.0/}{This work is licensed under a Creative Commons Attribution International 4.0 License.}
  \end{minipage}
  \vspace{5pt}
}
\makeatother

\begin{document}

\title{CKNN: Cleansed k-Nearest Neighbor for Unsupervised Video Anomaly Detection}



\author{Jihun Yi}
\orcid{0000-0001-5762-6643}
\affiliation{%
  \institution{Seoul National University}
  \city{Seoul}  
  \country{South Korea}
}
\email{t080205@snu.ac.kr}

\author{Sungroh Yoon}
\orcid{0000-0002-2367-197X}
\affiliation{%
  \institution{Seoul National University}
  \city{Seoul}  
  \country{South Korea}
}
\email{sryoon@snu.ac.kr}
\authornote{Correspondence to: Sungroh Yoon (\href{mailto:sryoon@snu.ac.kr}{sryoon@snu.ac.kr}).}









\renewcommand{\shortauthors}{Jihun Yi and Sungroh Yoon}


\newcommand{\xvec}{\mathbf{x}}
\newcommand{\Xvec}{\mathbf{X}}
\newcommand{\pvec}{\mathbf{p}}
\newcommand{\Atheta}{\mathcal{A}_{\theta}}
\newcommand{\mmap}{\mathcal{M}}
\newcommand{\Loss}{\mathcal{L}}

\newcommand{\textblue}{\textcolor{blue}}
\newcommand{\textred}{\textcolor{red}}
\newcommand{\xmark}{\text{\ding{55}}}
\newcommand{\cmark}{\text{\ding{51}}}

\newcommand{\etal}{\emph{et al.}}
\newcommand{\cond}{\,|\,}
\newcommand{\ie}{\textit{i}.\textit{e}.}
\newcommand{\eg}{\textit{e}.\textit{g}.}
\newcommand{\vs}{\textit{v}.\textit{s}.~}
\begin{abstract}
In this paper, we address the problem of unsupervised video anomaly detection (UVAD).
The task aims to detect abnormal events in test video using unlabeled videos as training data.
The presence of anomalies in the training data poses a significant challenge in this task, particularly because they form clusters in the feature space.
We refer to this property as the ``Anomaly Cluster'' issue.
The condensed nature of these anomalies makes it difficult to distinguish between normal and abnormal data in the training set.
Consequently, training conventional anomaly detection techniques using an unlabeled dataset often leads to sub-optimal results.
To tackle this difficulty, we propose a new method called Cleansed k-Nearest Neighbor (CKNN), which explicitly filters out the Anomaly Clusters by cleansing the training dataset.
Following the k-nearest neighbor algorithm in the feature space provides powerful anomaly detection capability.
Although the identified Anomaly Cluster issue presents a significant challenge to applying k-nearest neighbor in UVAD, our proposed cleansing scheme effectively addresses this problem.
We evaluate the proposed method on various benchmark datasets and demonstrate that CKNN outperforms the previous state-of-the-art UVAD method by up to 8.5\% (from 82.0 to 89.0) in terms of AUROC.
Moreover, we emphasize that the performance of the proposed method is comparable to that of the state-of-the-art method trained using \textit{anomaly-free} data.

\end{abstract}



\maketitle

\section{Introduction}
Increasing social demand for safety has led to a significant growth in the number of surveillance cameras.
While the use of security cameras is proliferating, human monitoring of these cameras is not scalable.
This has resulted in a need for computer vision-based video anomaly detection (VAD).
High-performing VAD methods have the potential to significantly enhance safety and security.

\begin{figure}
    \centering
    \includegraphics[width=\linewidth]{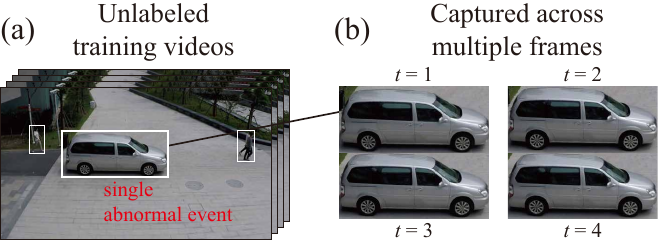}
    \vspace{-10pt}
    \caption{Anomaly Cluster issue. (a) UVAD methods use unlabeled videos that contain some anomalies as training data. (b) In those videos, each abnormal event is captured in multiple frames. The captured abnormal objects in each frame are similar to each other, and they form a cluster in feature space, as normal objects do. We refer to this as the \textbf{``Anomaly Cluster''} issue.}
    \label{fig:anomalycluster}
    \vspace{-10pt}
\end{figure}

The majority of previous research on VAD~\cite{future2018cvpr,memae2019iccv,appmot2021aaai,mnad2020cvpr,ssmt2021cvpr,jigsaw2022eccv,flowguided2021iccv,objae2019cvpr,pretexts2023iccv,eval2023cvpr} is based on anomaly-free training videos.
The methods learn normality from the anomaly-free training dataset and detect if test data deviate far from the training data.
We refer to this type of task as one-class VAD (OCVAD) in this paper.
Preparing anomaly-free data for OCVAD requires intensive human inspection, making it a costly process.
By contrast, unsupervised video anomaly detection (UVAD) requires a much weaker assumption: the \textit{majority} instead of \textit{all} of the training data needs to be normal.
In UVAD, we train a VAD model using a training dataset that may contain a few anomalies.
This relaxed assumption makes UVAD more applicable to the real-world problem because minimal human inspection is needed for UVAD data preparation.

For the same reason, the unsupervised scenario is gaining attention~\cite{soft2022neurips} in image anomaly detection (IAD). 
Anomalies in images are not necessarily similar to each other~\cite{deepsad}, and they tend to lie in low-density regions.
This property makes them easier to separate compared to those in videos.
As a result, the state-of-the-art one-class IAD method~\cite{patchcore2022cvpr} faces a minor performance degradation in the unsupervised scenario~\cite{soft2022neurips} (\eg, AUROC drops from 99.0 to 98.4).
However, unlike images, the performance degradation in videos is significant~\cite{ssmt2021cvpr,jigsaw2022eccv} because UVAD presents a unique challenge: the anomalies in videos form clusters.
This is because, in a video, a single abnormal event is recorded across multiple frames.
We provide examples in Fig.~\ref{fig:anomalycluster}.
In Fig.~\ref{fig:anomalycluster}(b), an abnormal event (a car appearing on a pavement) is recorded in multiple consecutive frames, and they look similar to each other.
As a result, their features appear nearby in the feature space.
For density estimators, such high-density anomaly clusters are indistinguishable from normal data, making it difficult to tell which one is normal.
This phenomenon, which we refer to as the ``Anomaly Cluster'' issue, is problematic in UVAD because VAD techniques may misjudge these clusters in unlabeled training data as normal, leading to sub-optimal detection performances.

Recently, the deployment of deep neural networks in UVAD has shown promising results~\cite{spr2022cvpr}, but the performance gap compared to OCVAD~\cite{jigsaw2022eccv} remains significant.
Since OCVAD scenario makes use of anomaly-free dataset, the detection performance of OCVAD methods~\cite{jigsaw2022eccv} has been considerably higher than that of UVAD methods~\cite{ordinal2020cvpr,spr2022cvpr}.

The aim of this research is to overcome the Anomaly Cluster issue and achieve UVAD performance that is comparable to that of OCVAD method.
To this end, we present cleansing scheme as a remedy to the Anomaly Cluster issue and demonstrate powerful anomaly detection performance by adopting nearest neighbor (NN) search, resulting in the proposal of Cleansed k-Nearest Neighbor (\textbf{CKNN}).
Our contributions can be summarized as follows.
\begin{itemize}
\item We propose a novel method, CKNN, that achieved state-of-the-art performance on four benchmark datasets in UVAD. The performance is even comparable to that of the state-of-the-art method in OCVAD~\cite{jigsaw2022eccv}, which learns with \textit{anomaly-free} data.
\item We identify, for the first time, the Anomaly Cluster issue in UVAD and develop a remedy that overcomes it. A direct application of NN search suffers from the issue and yields low performance, but our remedy significantly improves performance and demonstrates that NN search is practical in the UVAD scenario.
\item We provide the results of sensitivity tests regarding the hyperparameters of CKNN, verifying the robustness of the proposed method. We also provide a guideline for selecting the hyperparameter values, and a running time analysis shows that CKNN has minor computational overhead for real-time application.
\end{itemize}

\section{Related work}

\subsection{Video anomaly detection}
VAD methods typically separate the information in a video into two categories: appearance and motion~\cite{appmot2021aaai,eval2023cvpr}.
Appearance pertains to the visual information in a single frame, while motion relates to movement across multiple frames.
These two types of information are utilized differently in training VAD models.

Autoencoders (AEs)~\cite{future2018cvpr,memae2019iccv,mnad2020cvpr} are commonly used to model the appearance of video frames.
AEs are trained to either reconstruct the original frame~\cite{memae2019iccv} or predict the near future~\cite{future2018cvpr,pretexts2023iccv}.
The underlying principle is that AEs trained on an anomaly-free dataset cannot reconstruct or predict future for frames that contain abnormal events.
Thus, a large difference between the input and its reconstruction indicates the abnormality.
As in IAD~\cite{patchcore2022cvpr}, using a pretrained feature extractor~\cite{attr2022arxiv} is another option to effectively extract visual features and learn the normality of videos.

Motion in video can be learned either explicitly or implicitly.
One explicit way to extract motion from frames is to use optical flow~\cite{flownet2}.
Off-the-shelf networks such as FlowNet~\cite{flownet} estimate optical flow between two consecutive frames.
The optical flow map can be used to assist AE training for frame prediction~\cite{future2018cvpr}, or as a target of reconstruction~\cite{appmot2019iccv,appmot2021aaai}.
Additionally, there are heuristic methods for featurizing optical flow~\cite{attr2022arxiv} into low-dimensional vectors.
Besides, motion can be implicitly learned by using multiple consecutive frames as input to a model~\cite{memae2019iccv}.

To further enhance the accuracy of anomaly detection, some work~\cite{objae2019cvpr} has proposed the adoption of object detectors.
Pretrained object detectors (\eg~YOLO~\cite{yolov3}) extract objects in every frame, and a spatio-temporal cube around the objects becomes the input to a VAD method.
This pre-processing effectively reduces the search space for anomalies.
The VAD module, following the object detector, estimates object-wise anomaly scores, which are aggregated to compute frame-wise scores.
Due to its effectiveness in improving detection performance, various recent state-of-the-art methods in OCVAD~\cite{ssmt2021cvpr,jigsaw2022eccv,pretexts2023iccv} and UVAD~\cite{spr2022cvpr} make use of object detectors.
Notably, many VAD techniques predominantly rely on pretrained networks (\eg, feature extractor, optical flow estimator, and object detector), and effectively leveraging their capability is essential for successful anomaly detection in VAD.

Apart from these, there are recent approaches that introduce concepts such as diffusion~\cite{diffusion2023iccv} or video event restoration~\cite{event2023cvpr}.
In addition, there has been work proposing the removal of smaller clusters from the feature clusters of training data~\cite{narrowed}.
This method aims to discard small outliers present in the OCVAD setup from the training data.
However, due to the previously mentioned Anomaly Cluster issue, its effectiveness diminishes in UVAD where anomalies form larger clusters.
Furthermore, as the decision to discard is made at the cluster level, individual data points are not considered, which leads to sub-optimal performance in UVAD.

\subsection{Unsupervised video anomaly detection}
UVAD is a VAD problem that involves training with unlabeled videos.
Differing slightly from UVAD, in the past, methods were proposed that make judgments on individual test videos without using any training videos at all.
Initial train-less VAD research utilized classifier-based methods to detect sudden frame changes~\cite{discriminative2016eccv}, identifying frames that classifiers easily separate as abnormal.
Later studies~\cite{unmasking2017iccv,classifier2018bmvc} applied the unmasking technique in VAD, training classifiers to differentiate between past and future frames by progressively removing features.
Abnormal frames are easily classified even after multiple feature removals.

More recently, methods utilizing unlabeled training videos have been proposed.
An ordinal regression-based approach~\cite{ordinal2020cvpr} in conjunction with self-training generates pseudo-labels for each frame and iteratively updates these labels.
The label-generating neural network becomes the final scoring function for anomaly detection.
Recently, LBR-SPR~\cite{spr2022cvpr} proposed a self-training-based approach to train AEs.
The authors iteratively removed training data with high reconstruction loss during training in a self-training fashion.

In contrast to this self-reinforcing approach, our method exhibits minimal reliance on randomness in its initial training phase.
Furthermore, our method can incorporate any techniques for sanitizing training data, offering a generalized cleansing scheme.

\subsection{k-NN in anomaly detection}
The k-nearest neighbor (k-NN) method is a long-standing algorithm in machine learning and anomaly detection~\cite{knn}.
In the field of anomaly detection~\cite{knn_tabular,patchcore2022cvpr}, the nearest neighbors of a query data are used for the calculation of its anomaly score based on the distance to the selected neighbors.
Specifically, for a query data $x_q$, and a search space $\mathbf{X}$, k-NN calculates its anomaly score as follows:

\begin{equation} \label{eq:knn}
    \mathrm{kNN}(x_q, \mathbf{X}, k) = \frac{1}{k} \sum_{x \in \mathcal{N}_k (\mathbf{X}, x_{q})} \left \| x_{q} - x \right \|_2,
\end{equation}
where $\mathcal{N}_k (\mathbf{X}, x_q)$ refers to the set of $k$ nearest neighbors of $x_q$ in $\mathbf{X}$.
In the field of IAD, k-NN has demonstrated promising results~\cite{dn22020,spade2020,patch2020accv,patchcore2022cvpr}, and most state-of-the-art methods~\cite{ucad,jeong2023winclip,hyun2024reconpatch} utilize k-NN as a tool for anomaly score calculation.
Compared to IAD, the adoption of k-NN in VAD~\cite{attr2022arxiv}, particularly in UVAD, remains relatively under-explored.
In this paper, we aim to leverage simple yet effective k-NN algorithm to address UVAD task.

\section{Motivation}

\subsection{Problem formulation}

We first formalize the UVAD problem and introduce notations.
Given $\mathbf{D} = \left\{ v_1, v_2, ..., v_N \right\}$, a set of videos, $v_i = \left\langle f^i_1, f^i_2, ..., f^i_{T_i} \right\rangle$ refers to a video of length $T_i$.
For each frame $f^{i}_t$, the corresponding ground truth label $y^{i}_t$ indicates whether the frame is normal ($y^{i}_t = 0$) or abnormal ($y^{i}_t = 1$).
OCVAD assumes $y^{i}_t = 0$ for all frames, while UVAD assumes that the corresponding label $y^{i}_t$ is not accessible.
For brevity, we omit the video index, $i$, from here on.
Using $\mathbf{D}$ as a training set, we train a scoring function ($\mathcal{A}$) that estimates the anomaly score ($a_t$) for each frame of a test video as follows.

\begin{equation} \label{eq:scoring_function}
    \mathcal{A}(v^{test}, t) = a_t.
\end{equation}

\begin{table}[t]
\centering
\caption{Comparison to the reconstruction loss-based state-of-the-art method~\cite{spr2022cvpr}.}  
\begin{tabular}{c | c c c c}
   \hline
    Method & & Scenario & AVE & STC \\
    \hline
\hline
\multirow{2}{*}{Recon-based~\cite{spr2022cvpr} }  & (\textred{a}) & OCVAD & 93.2 &  82.4    \\    
                              & (\textred{b}) & UVAD & 93.0 & 82.0 \\  
\hline
\multirow{2}{*}{ k-NN } & (\textred{c}) & OCVAD &  94.4 & 90.4  \\
& (\textred{d})& UVAD & 73.5 & 76.3  \\
\cdashline{1-5}
\makecell{\textbf{k-NN} \\ \textbf{+ Cleansing}} & (\textred{e}) & UVAD & \textbf{94.1} & \textbf{89.0}  \\
\hline
\end{tabular}%
\label{table:motivation}
\end{table}
To train the scoring function described in Eq.~\ref{eq:scoring_function}, and to capitalize on the benefits of object detectors as demonstrated in prior research~\cite{objae2019cvpr,ssmt2021cvpr,spr2022cvpr}, we begin by detecting objects in the input frames.
For each input frame, $f_t$, a pre-trained object detector~\cite{maskrcnn} identifies a collection of bounding boxes,
\begin{equation} \label{eq:object_detection}
    \mathcal{B}_t = \left\{ b_{(t, 1)}, b_{(t, 2)}, ..., b_{(t, M_t)} \right\},
\end{equation}
where $M_t$ represents the number of objects detected in the frame $f_t$.
We aim to build object-based methods to assess abnormalities at the individual object level, subsequently aggregating these assessments to evaluate the overall frame abnormality.
In this approach, we employ a k-NN technique to contrast the objects in the test frames with those in the training frames within a specified feature space.

\subsection{Performance degradation in UVAD} \label{sec:uvad_degradation}

Unlike OCVAD, the UVAD scenario involves training data that contains anomalies, and methods that do not account for this can experience significant performance degradation.
This is detailed further in Sec~\ref{sec:ocvad_to_uvad}, where the application of OCVAD methods to the UVAD scenario results in substantial performance drops.
A previous state-of-the-art approach~\cite{spr2022cvpr} for UVAD employs reconstruction loss for anomaly scoring.
The authors demonstrated that the reconstruction loss using AEs inherently exhibits some robustness to anomalies in the training data, a feature they named the Normality Advantage.
Thanks to this robustness and additional techniques such as self-training, their LBR-SPR~\cite{spr2022cvpr} model shows negligible differences in performance between OCVAD and UVAD scenarios (Table~\ref{table:motivation} (\textred{a}) \vs (\textred{b})).
The OCVAD scenario involves training with anomaly-free data, while the UVAD scenario includes anomalies within the training data.
Table~\ref{table:motivation} displays the performance in OCVAD and UVAD scenario on the Avenue (AVE)~\cite{avenue} and ShanghaiTech Campus (STC)~\cite{stc} datasets, quantified using the commonly used AUROC metric.

In the OCVAD scenario, where anomaly-free training data is accessible, k-NN outperforms the reconstruction loss-based detection baseline (Table~\ref{table:motivation} (\textred{c}) \vs (\textred{a})).
This trend is also observed in the image anomaly detection domain, where most state-of-the-art methods~\cite{spade2020,patch2020accv,dn22020,ucad,jeong2023winclip,hyun2024reconpatch} utilize k-NN based feature comparison and outperforms reconstruction-based methods~\cite{RnD,akcay2019ganomaly,ocgan2019cvpr}.
Despite its strength over recon-based method in the OCVAD scenario, k-NN lacks robustness to anomalies, leading to a significant performance drop in the UVAD setup (Table~\ref{table:motivation} (\textred{c}) \vs (\textred{d})).
We have identified that the cause of this issue lies in the Anomaly Cluster issue, which will be introduced in the following section.
By addressing the anomaly clusters through object cleansing, we achieve both robustness and high performance in k-NN (Table~\ref{table:motivation} (\textred{c}) \vs (\textred{e})), striking a better balance between performance and robustness compared to the recon-based baseline~\cite{spr2022cvpr}(Table~\ref{table:motivation} (\textred{e}) \vs (\textred{b})).

\begin{figure}[t]
\centering
\begin{minipage}{0.42\linewidth}
\centering
\includegraphics[width=0.75\linewidth]{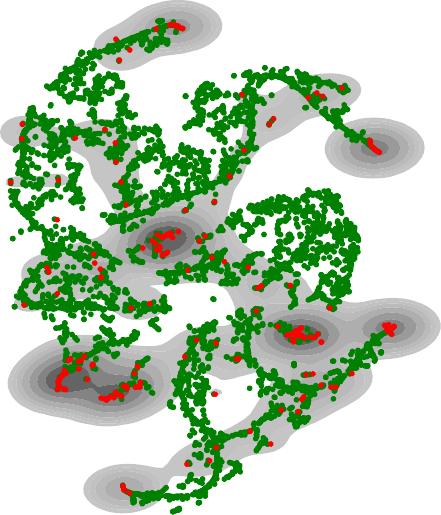}
\centering
\vspace{-8pt}
\caption{t-SNE~\cite{tsne} of training object features.}
\label{fig:tsne_ori}
\end{minipage}
\begin{minipage}{0.1\linewidth}
\hfill
\end{minipage}
\begin{minipage}{0.45\linewidth}
\centering
\includegraphics[width=\linewidth]{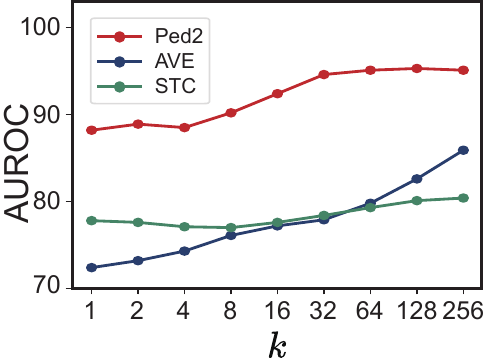} 
\centering   
\caption{The effect of increasing $k$ in k-NN.}
\label{fig:varying_k_ave}
\end{minipage}
\end{figure}

\begin{figure*}[t]
    \centering
    \includegraphics[width=\textwidth]{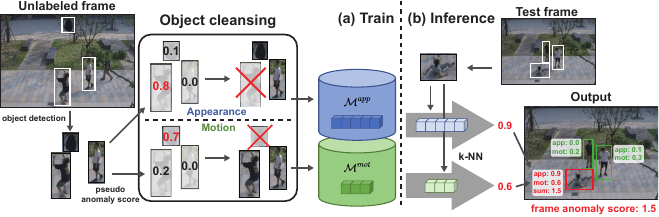}
    \vspace{-10pt}
    \caption{
    Overall flow of the proposed method.
    (a) At train phase, an object detector extracts objects in input frames, and CKNN cleanses the objects using two types of pseudo-anomaly scores: appearance and motion. The cleansed objects are stored as features in the two types of feature banks. (b) For inference, CKNN estimates appearance and motion anomaly scores for each object in the test frames using k-NN algorithm. The sum of the two scores is the anomaly score of each object, and the maximum value in a test frame becomes its final anomaly score.
    }
    \vspace{-10pt}
    \label{fig:overall}
\end{figure*}

\subsection{Anomaly clusters}

As demonstrated in Fig.~\ref{fig:anomalycluster}, anomalies within videos tend to form clusters.
Fig.~\ref{fig:tsne_ori} presents a t-SNE visualization~\cite{tsne} of the optical flow features~\cite{attr2022arxiv} from the training data of ShanghaiTech Campus dataset, where abnormal objects indicated in red cluster together as densely as the normal ones marked in green.
This visualization reaffirms the clustering behavior of anomalies.

A significant challenge in applying k-NN to the UVAD problem is the presence of anomalies within the training data.
When anomalies are included among the nearest neighbors (as denoted by $\mathcal{N}_k (\mathbf{X}, x_{q})$ in Eq.~\ref{eq:knn}), they can disrupt the accurate estimation of abnormality.
Increasing the value of $k$ might reduce the proportion of anomalies in $\mathcal{N}_k (\mathbf{X}, x_{q})$, thus enhancing detection performance.
This improvement is observed in Fig.~\ref{fig:varying_k_ave}, which shows increased anomaly detection performance across three benchmark datasets.
However, since anomalies form clusters, to fully mitigate their influence, $k$ needs to be significantly larger than the size of the anomaly clusters.
This approach is impractical due to memory constraints and can dilute the precision of the k-NN method. 

\begin{table}[t]
\centering
\captionof{table}{The effect of temporal object merging.}
\begin{tabular}{c c}\toprule  
 & AUROC \\\midrule
No merging & 76.3 \\ 
Merged 25\% &  78.4 \\
Merged 80\% &  81.7 \\ \midrule
\textbf{Ours} &  \textbf{89.0} \\  \bottomrule
\end{tabular}
\label{table:downsample}    
\end{table}

Another straightforward method involves merging overlapping boxes along the time axis.
This can merge similar anomalous bounding boxes into one, thereby reducing the density and size of anomaly clusters.
The anomaly detection performances of this rudimentary technique are presented in Table~\ref{table:downsample}.
As the percentage of merged boxes increases from 0\% (No merging) to 80\%, there is an improvement in the anomaly detection performance, but it plateaus at a sub-optimal level.
Given the less-than-ideal results of these two simple approaches, we introduce an explicit object cleansing scheme in the subsequent section.

\section{Methodology}
In this section, we describe the proposed method, CKNN.
Our method uses k-NN as the backbone, with the aim of addressing the significant challenge of anomaly clusters in UVAD through a technique called object cleansing.
An overview of the proposed method is provided in Fig.~\ref{fig:overall}.

\subsection{Object cleansing} \label{sec:denoising}
In this section, we proactively reduce abnormal objects from the training dataset.
Object cleansing involves estimating a \textit{pseudo-anomaly score} for each object and discarding those with high scores.
As a result, the remaining set of objects is anticipated to have fewer anomalies and anomaly clusters.
We conduct object cleansing in two dimensions: appearance and motion.
This produces two sets of objects, denoted as $\mathcal{D}^{app}$ and $\mathcal{D}^{mot}$.
Various experimental analyses provided in Sec~\ref{sec:exp} demonstrate that performing object cleansing along these two dimensions is highly effective.

The pseudo-anomaly score estimators can be any scalar functions, and we denote them as $f^{app}$ and $f^{mot}$.
The two corresponding pseudo-anomaly scores for an object $b_{(t, m)}$, labeled as $\hat{s}^{app}_{(t, m)}$ and $\hat{s}^{mot}_{(t, m)}$, are calculated as in Eq.~\ref{eq:pseudo_scorer}.
\begin{equation} \label{eq:pseudo_scorer}
    \hat{s}^{app}_{(t, m)} =  f^{app}(b_{(t,m)}),~\hat{s}^{mot}_{(t, m)} =  f^{mot}(b_{(t,m)}).
\end{equation}

After computing the scores, we exclude objects with scores in top-$\tau$\% range.
The remaining objects are then stored in a feature bank using feature extractors, $\phi^{app}$ and $\phi^{mot}$, as described in Eq.~\ref{eq:featurize}.
Each removal results in purified memory banks, denoted as $\mathcal{D}^{app}$ and $\mathcal{D}^{mot}$, respectively.

\begin{equation} \label{eq:featurize}
\begin{aligned}
   \mathcal{D}^{*} & = \left\{ \phi^{*}(b_{(t,m)}) \cond \hat{s}^{*}_{(t, m)} < \texttt{top}^{*} (\tau) \right\},  \\
\end{aligned}
\end{equation}
where $*$ denotes either app or mot for the appearance and motion, respectively.
In this context, $\tau$ is a hyperparameter dictating the extent of cleansing; a smaller $\tau$ removes fewer objects, while a larger $\tau$ leads to more aggressive cleansing.

\subsection {Feature bank compression} \label{sec:normality}
Optionally, we can compress the feature bank through random sampling to reduce memory consumption.
For both the appearance and motion memory banks, we randomly select $p$\% of the features, as expressed in Eq.~\ref{eq:memorybank}.

\begin{equation} \label{eq:memorybank}
    \mathcal{M}^{app} = \texttt{rand}(\mathcal{D}^{app}, p), \mathcal{M}^{mot} = \texttt{rand}(\mathcal{D}^{mot}, p). \\
\end{equation}

Experiments using coreset selection~\cite{agarwal2005geometric} for more effective compression did not show a significant difference in performance.
The related results are provided in Sec~\ref{sec:compression}.

\subsection{Inference}
At test time, we examine each frame of a given test video.
For each frame, we detect objects in the frame and perform object-wise inspection based on its similarity to the training objects using k-NN algorithm (Eq.~\ref{eq:knn}).
Each object, $b^{test}$, is described by a pair of two features, $\phi^{app}(b^{test})$ and $\phi^{mot}(b^{test})$, and $\mathcal{M}^{app}$ and $\mathcal{M}^{mot}$ are their respective k-NN search spaces, as shown in Eq.~\ref{eq:anomalyscore_object}.

\begin{table}[t]
     \centering
     \setlength\tabcolsep{4pt}
  \caption{Statistics of the datasets. Objects$^{\dagger}$ refer to the bounding boxes detected by Mask R-CNN~\cite{maskrcnn}.}
  \vspace{-15pt}
   \begin{tabular}{l r r r r r}
   \\
   \hline
         & & \multicolumn{1}{c}{Ped2} & \multicolumn{1}{c}{AVE} & \multicolumn{1}{c}{STC} & \multicolumn{1}{c}{UBn} \\     
    \hline    
    \hline
    \\[-0.9em]    
    \multirow{2}{*}{\# videos} & train & 16 & 16 & 330 & 268 \\
     & test & 12 & 21 & 107 & 211 \\
    \hline
    \multirow{2}{*}{\makecell{Contains \\ anomaly}} & train & \multicolumn{1}{c}{\xmark} & \multicolumn{1}{c}{\xmark} & \multicolumn{1}{c}{\xmark} & \multicolumn{1}{c}{\cmark}  \\
    & test & \multicolumn{1}{c}{\cmark} & \multicolumn{1}{c}{\cmark} & \multicolumn{1}{c}{\cmark} & \multicolumn{1}{c}{\cmark} \\       
    \hline   
    \multirow{2}{*}{\# objects$^{\dagger}$} & train & 29,970 & 113,751 & 1,140,631 & 650,937 \\
                                & test  & 33,133 &  96,147 &  174,112 & 488,964\\
    \hline
    \multirow{2}{*}{\makecell{\% abnormal \\ objects$^{\dagger}$ ($r$)}}  & train & 0.0\% & 0.0\% & 0.0\% & 16.7\% \\
      & test & 10.3\% & 4.1\% & 12.6\% & 21.3\% \\
    \hline
    \end{tabular}%
  \centering
  \label{table:datasets}
\end{table}

\begin{equation} \label{eq:anomalyscore_object}
\begin{aligned}
    s^{app} & = \mathrm{kNN}(\phi^{app}(b^{test}), \mathcal{M}^{app}, k), \\
    s^{mot} & = \mathrm{kNN}(\phi^{mot}(b^{test}), \mathcal{M}^{mot}, k),
\end{aligned}
\end{equation}
where $s^{app}$ and $s^{mot}$ are object-wise appearance and motion anomaly scores, respectively.
In a test video frame, an object frame should be detected as anomaly if its appearance \textit{or} motion is abnormal.
Therefore, we sum the two scores for the overall anomaly score of the object.
Before the addition, we normalize each score using their values evaluated on the corresponding training sets.

We aggregate the object-level anomaly scores into frame-level scores by taking the maximum values and apply Gaussian smoothing for temporal smoothness.
As a result, video frames with high anomaly scores are regarded as abnormal.
The pseudocode and the implementation details of the proposed method are provided in the supplementary materials, and the code is available online\footnote{\url{https://github.com/nuclearboy95/Anomaly-Detection-CKNN-PyTorch}}.

\section{Experiments} \label{sec:exp}
\subsection{Experimental setup}
\subsubsection{Datasets}
To empirically evaluate the proposed method, we applied it to several benchmark datasets in VAD: UCSD Ped2 (Ped2) \cite{ped2}, Avenue (AVE)~\cite{avenue}, and ShanghaiTech Campus (STC)~\cite{stc}, which are commonly used in UVAD~\cite{spr2022cvpr,ordinal2020cvpr}.
These datasets consist of videos recorded on streets, making them well-suited for developing anomaly detection models for surveillance system.
Normal scenes in the videos include people walking down the streets, while abnormal events include unexpected objects (\eg~bicycle) and unexpected actions (\eg~people falling down).
Additionally, we utilized the recently proposed UBnormal (UBn) dataset~\cite{ubnormal}.
Unlike the previous three datasets, its train split includes anomalies, making it effective for UVAD experiments.
A few statistics about the datasets are provided in Table~\ref{table:datasets}.

\begin{figure}[t]
    \centering
    \includegraphics[width=0.6\linewidth]{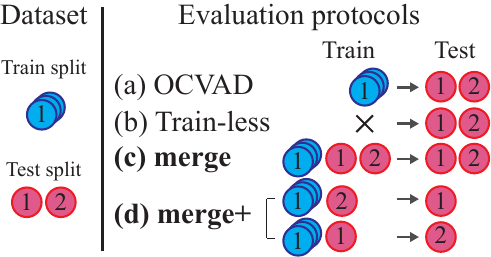}
    \vspace{-8pt}
    \caption{UVAD evaluation protocols using datasets having anomaly-free train splits.}    
    \label{fig:scenario}
\end{figure}

\subsubsection{Evaluation scenarios}
The datasets provide train and test splits.
The UBn dataset includes anomalies in its train split, thus allowing for UVAD evaluation using the defined train/test split.
However, the train splits of the other three datasets contain no anomalies, necessitating a reorganization of the train and test data for UVAD evaluation.
In prior studies of UVAD~\cite{spr2022cvpr,ordinal2020cvpr}, two types of evaluation protocols have been employed: \textit{partial} mode and \textit{merge} mode.
Both modes use the test splits as test dataset, but they have differences in the composition of training data.
To add anomalies to the training data, they utilized the test splits at train time; in \textit{partial} mode, test splits alone are used in training, while \textit{merge} mode utilizes both train and test splits in training.
Note that the labels are never used at train time in both modes.
The schematic description of the merge mode is provided in Fig.~\ref{fig:scenario}(c).

The two evaluation protocols mentioned above, while not using test split labels, have the issue of potential overlap where the same video frames could be used during both training and testing.
To avoid this problem, we propose novel \textit{merge+} scenario, which ensures individual test videos used for evaluation are excluded from the training set, as depicted in Fig.~\ref{fig:scenario}(d). 
We believe that this evaluation protocol can be considered more cautious as it prevents data overlap between training and testing set.
For continuity with previous research, we report the results of both the partial and merge modes, as well as the merge+ mode protocol for the Ped2, AVE, and STC datasets, while using the train/test splits as-is for the UBn dataset.

At test time, we compute the frame-wise anomaly score and measure anomaly detection performance using AUROC (area under receiver operating characteristic curve) for each video, and mean of the AUROCs becomes the final metric.

\begin{figure*}[t]
    \centering    
    \includegraphics[width=0.9\textwidth]{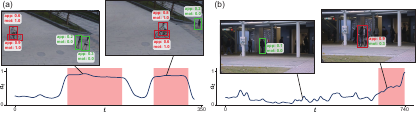}
    \vspace{-10pt}
    \caption{CKNN output for test videos in (a) STC and (b) AVE datasets. Estimated anomaly scores for each frame is plotted at the bottom, and red shades indicate that the corresponding frames are labeled as anomalies in the ground truth annotations. For four exemplar video frames, detected bounding boxes and their appearance and motion anomaly scores are depicted. Note that the frames are zoomed for better visualization.}
    \vspace{-10pt}
    \label{fig:example}
\end{figure*}

\subsubsection{Implementations}

Object cleansing requires various components, including $\phi^{app}$, $\phi^{mot}$, $f^{app}$, and $f^{mot}$.
We have experimented with different implementations for each component, all of which have successfully performed object cleansing and demonstrated effectiveness.
Detailed results of these experiments are provided in the subsequent ablation study in Sec~\ref{sec:ablation1}.
Throughout the experimental section, unless otherwise specified, we adopted the following implementations.
For $\phi^{app}$ and $\phi^{mot}$, we used CLIP~\cite{clip} encoder and the hand-crafted optical flow feature extractor~\cite{attr2022arxiv}, respectively.
For $f^{app}$, we used AE reconstruction loss, \ie, $f^{app}(b)=\left \| b - \texttt{AE} ( b) \right \|_2$, where \texttt{AE} indicates an AE trained on whole objects in training dataset.
For $f^{mot}$, we used Gaussian Mixture Model (GMM) trained using the optical flow features~\cite{attr2022arxiv} of  whole objects in training dataset.
Throughout the experiments, we used hyperparameters values of $k=4$ for k-NN and $n=8$ for GMM, where $n$ is the number of components of GMM.
For $\tau$, we used 15 for Ped2 and AVE datasets and 25 for STC and UBn dataset.
For more details, please refer to the supplementary materials.

\begin{table}[t]
\centering
\setlength\tabcolsep{5pt}
   \caption{Comparison of anomaly detection performances with the state-of-the-art methods using AUROC. In UVAD scenario, P, M, and M+ refer to the partial, merge, and merge+ modes, respectively. Bold fonts represent the best  performances in each evaluation protocol.}
\begin{tabular}{c l | c c c c | c}
\hline
      & Methods & & Ped2 & AVE & STC & UBn \\
\hline    
\hline
    
\parbox[t]{3.5mm}{\multirow{4}{*}{\rotatebox[origin=c]{90}{OCVAD}}}

& $\text{SSMT}$~\cite{ssmt2021cvpr}  & &  99.8 & 92.8 & 90.2 & \\
& $\text{Jigsaw}$~\cite{jigsaw2022eccv}  & &  \textbf{99.9} & \textbf{93.0} & \textbf{90.6} & \\
& $\text{Pretexts}$~\cite{pretexts2023iccv}  & &  97.6 & 90.9 & 78.8 & \\
& $\text{USTN-DSC}$~\cite{event2023cvpr}  & &  98.1 & 89.9 & 73.8 & \\

\hline
\parbox[t]{3.5mm}{\multirow{7}{*}{\rotatebox[origin=c]{90}{UVAD}}} 
& $\text{Ordinal}$~\cite{ordinal2020cvpr} & M  & 83.2 & & & \\  \cdashline{2-7}

& \multirow{3}{*}{$\text{LBR-SPR}$~\cite{spr2022cvpr}} & P  & 95.7 & 87.7 &  83.0 &  \\
 & & M  & 98.1 & 93.0 & 82.0 & 77.6 \\
 & & M+  & 94.0 & 83.2 & 81.0 & \\ \cdashline{2-7}
& \multirow{3}{*}{\textbf{CKNN (Ours)}}   & P & \textbf{99.8} &  \textbf{94.0} &  \textbf{88.7} &   \\    
                             & & M & \textbf{99.7} & \textbf{94.1} & \textbf{89.0} & \textbf{83.5}\\    
                             & & M+  & \textbf{99.6} & \textbf{95.4} & \textbf{88.7} & \\

\hline
\end{tabular}%
     \centering
  \label{table:sota}
\vspace{-10pt}
\end{table}

\subsection{Outputs of CKNN}
In Fig.~\ref{fig:example}, we present the examples of the output of the proposed method, CKNN.
The bottom plots show the estimated anomaly score for each frame.
Video frames with high anomaly scores are likely to contain abnormal events.
We can observe that CKNN successfully assigns high anomaly scores to abnormal frames, which are depicted as red shades.

In Fig.~\ref{fig:example}(a), the anomaly score rapidly rises as a bicycle (anomaly) appears.
The video frames show that both the bicycle and the person riding it receive high appearance and motion anomaly scores because they look and move abnormally.
In Fig.~\ref{fig:example}(b), a person slowly dragging a bike receives high appearance anomaly score due to the bike, along with low motion score as it moves as slowly as normal walking person.
More examples are provided in the supplementary materials.
Qualitatively, the appearance and motion anomaly scores for each object align with human intuition.
The accurate detection of object-wise anomalies contributes to the high quantitative evaluation results discussed in the following section.

\begin{figure}[t]
    \centering
    \includegraphics[width=0.9\linewidth]{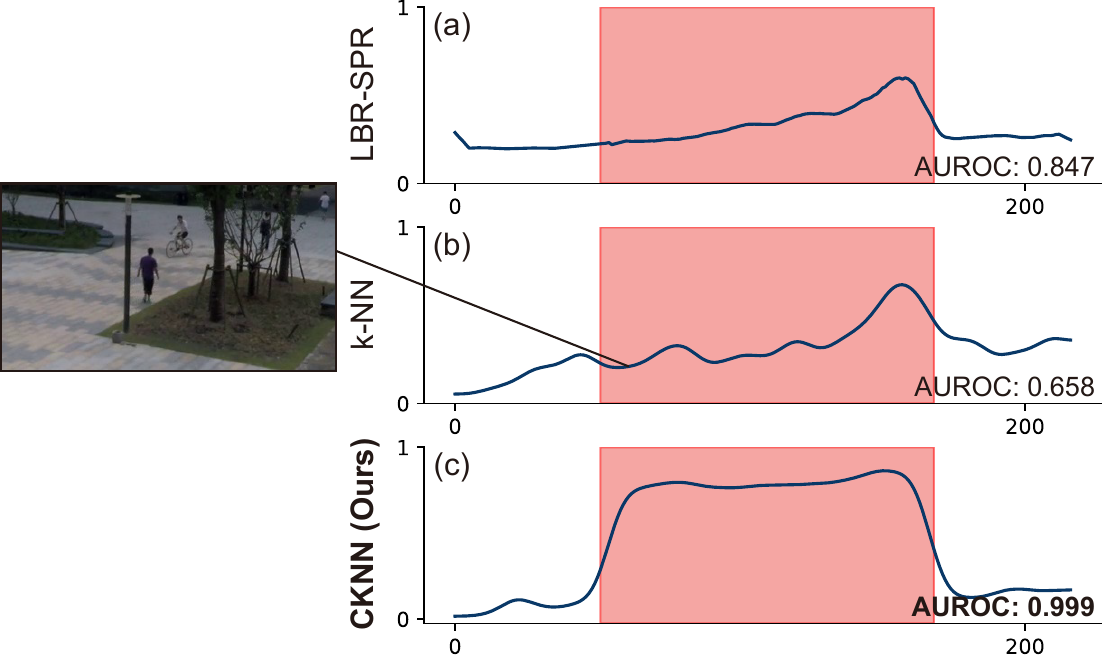}    
    \vspace{-5pt}
    \caption{Comparison of the anomaly scores estimated using the reconstruction-based LBR-SPR~\cite{spr2022cvpr}, k-NN, and CKNN.}
    \label{fig:compare_scores}    
\end{figure}

\begin{figure*}[t]
    \centering
    \includegraphics[width=\linewidth]{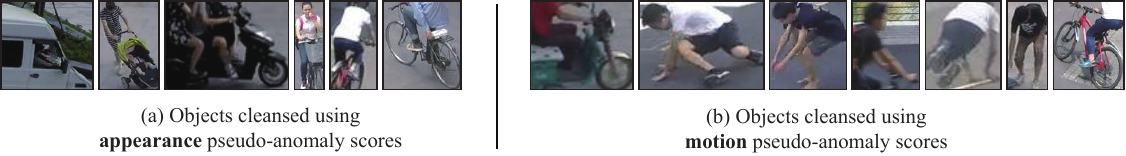}
    \vspace{-8pt}
    \caption{Examples of objects removed by the object cleansing procedure.}    
    \label{fig:cleansed_objects}
\end{figure*}

\subsection{Comparison to the other methods}
To demonstrate the efficacy of the proposed method, we conducted experiments and compare its UVAD performance with that of the state-of-the-art (SOTA) methods in Table~\ref{table:sota}.

Our results show that the proposed method outperforms all previous UVAD methods~\cite{ordinal2020cvpr,spr2022cvpr} in all evaluation modes and datasets.
It is worth noting that the detection performance of the proposed method is even comparable to that of the OCVAD SOTA method~\cite{jigsaw2022eccv}, which learns with anomaly-free data.
This indicates that CKNN is robust to the anomalies in the training datasets. This robustness is significantly enhanced by our object cleansing technique, as detailed in the following analysis.
The results of LBR-SPR~\cite{spr2022cvpr} are gained from running the official code release\footnote{\url{https://github.com/yuguangnudt/LBR_SPR}} by the authors.

Fig.~\ref{fig:compare_scores} presents the anomaly detection results for an STC video using LBR-SPR~\cite{spr2022cvpr}, k-NN, and the proposed method (CKNN).
The reconstruction-based LBR-SPR in Fig.~\ref{fig:compare_scores}(a) shows insensitive detection results, as it does not immediately alert to the anomaly score despite the appearance of an abnormal event, \ie, a bicycle.
As Fig.~\ref{fig:compare_scores}(b) depicts, k-NN also fails to accurately identify abnormal events as anomalies.
This is because the anomalies in the training data are included in Eq.~\ref{eq:knn} computation, leading to the assignment of low anomaly scores to the anomaly.
In contrast, CKNN exhibits an immediate response upon the occurrence of an abnormal event, outputting highly accurate anomaly detection results.

\subsection{The effect of object cleansing}
Object cleansing is a key technique that makes CKNN robust to the issue of anomaly clusters.
In this section, we analyze the effectiveness of object cleansing from multiple perspectives.

\begin{figure}[t]
    \centering
    \includegraphics[width=\linewidth]{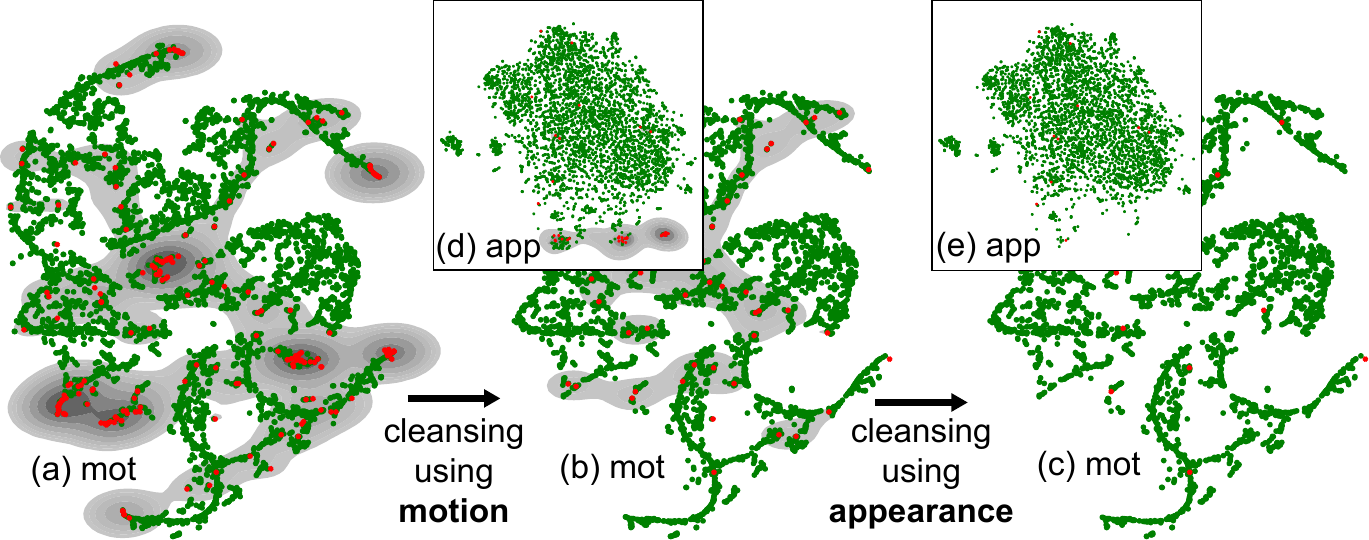}
    \vspace{-10pt}
    \caption{t-SNE~\cite{tsne} of object features when object cleansing undergoes. (a-c) motion features and (d-e) appearance features. Red (abnormal) and green (normal) indicate normality.}
    \label{fig:tsne}
    \vspace{-10pt}
\end{figure}

\subsubsection{Examples of cleansed objects}

Fig.~\ref{fig:cleansed_objects} showcases uncurated collections of objects from the training videos that received the top 1\% of pseudo-anomaly scores and were subsequently removed during the object cleansing process.
Fig.~\ref{fig:cleansed_objects}(a) shows objects that received high \textit{appearance} pseudo-anomaly scores, primarily consisting of objects with abnormal appearances such as cars, strollers, and bicycles.
Fig.~\ref{fig:cleansed_objects}(b) presents objects that received high \textit{motion} pseudo-anomaly scores, which largely include individuals displaying abnormal behaviors such as falling or jumping.
Motorcycles and bicycles are also included, likely due to their movements being significantly different from those of a normally walking person.
These examples illustrate that the object cleansing appropriately filters the objects in terms of both appearance and motion.

\begin{figure}[t]
    \centering
    \includegraphics[width=0.9\linewidth]{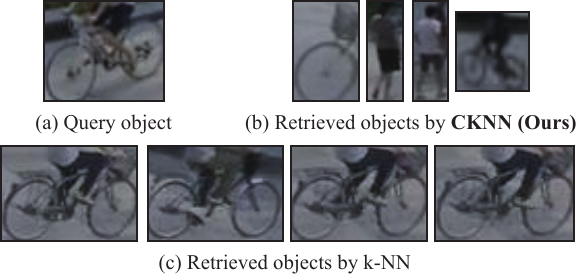}    
    \caption{For (a) a query object image, retrieved objects by proposed (b) CKNN and (c) k-NN are presented.}
    \label{fig:retrieved_objects}    
\end{figure}

\subsubsection{Feature visualization} \label{sec:tsne}

Fig.~\ref{fig:tsne} shows t-SNE~\cite{tsne} visualization of 6,000 sampled object features of STC training data for merge mode, which is also presented in Fig.~\ref{fig:tsne_ori}.
First, as Fig.~\ref{fig:tsne}(a) shows, anomalies in training data are as densely clustered as normal data, making them similar to normal clusters, exhibiting the Anomaly Cluster issue.
Object cleansing using motion (\textit{mot}) pseudo-anomaly score filters out major clusters in \textit{mot} feature space, but a few anomalies remain in Fig.~\ref{fig:tsne}(b).
It turns out that most of the remaining anomalies correspond to clusters in the appearance (\textit{app}) feature space (d), which are filtered out using \textit{app} object cleansing.
As seen above, object cleansing using \textit{app} and \textit{mot} pseudo-anomaly scores has complimentary effect, successfully filtering anomaly clusters and making training data more appropriate for k-NN.
Quantitative evidence behind their complimentary effect is provided in Sec~\ref{sec:app_and_mot}.

\subsubsection{Examples of retrieved objects in NN search}
In Fig.~\ref{fig:retrieved_objects}, we present the nearest neighbors of a query image, obtained by CKNN and ordinary k-NN from the training videos.
As a dataset-defined anomaly, bicycles should not be frequently retrieved in NN searches, even if they are present in the training videos to ensure higher accuracy in anomaly score estimation.
As illustrated in Fig.~\ref{fig:retrieved_objects}(c), for the image of anomaly `bicycle', k-NN returns a similar set of bicycle images.
These are close in distance from the query image, thus Eq.~\ref{eq:knn} using these neighbors assigns low anomaly score to the abnormal query image.
This is exemplified in Fig.~\ref{fig:compare_scores}(b).
Due to the Anomaly Cluster issue, increasing $k$ further only results in retrieving more similar bicycle images, and the anomaly score estimation remains inaccurate.
With the proposed object cleansing method, as shown in Fig.~\ref{fig:retrieved_objects}(b), these clusters are removed, and as a result, semantically dissimilar objects are retrieved in NN search.
They are distant from the query image the feature space, successfully assigning higher anomaly score to the query image.

\subsubsection{Effectiveness in UVAD}
\begin{figure}[t]
    \centering
    \includegraphics[width=\linewidth]{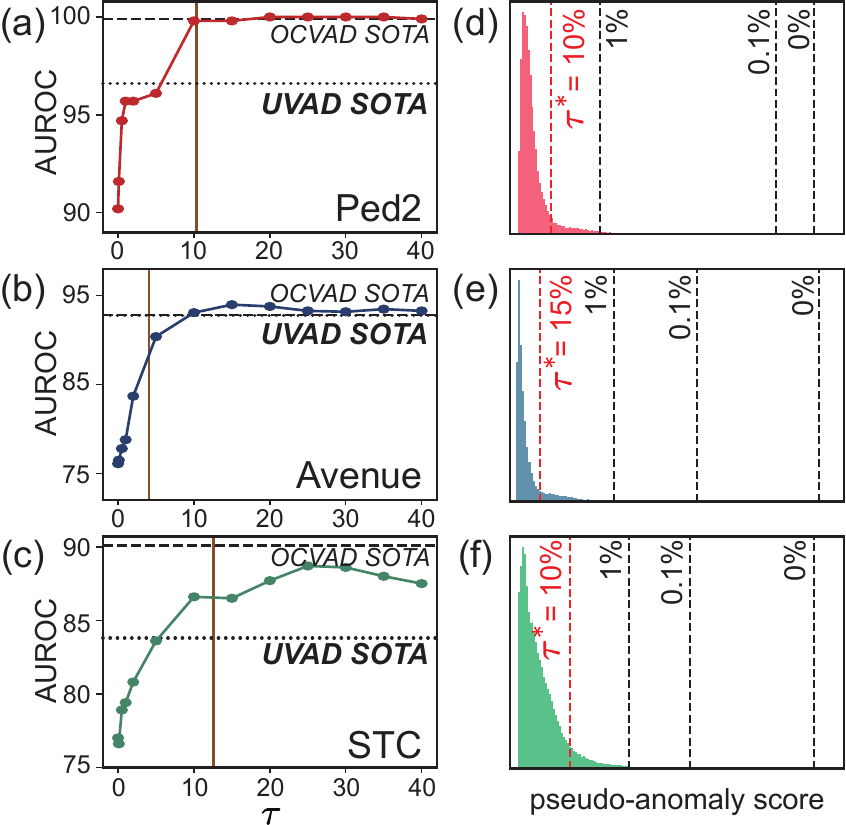}
    \vspace{-20pt}
    \caption{(a-c) CKNN performances for varying $\tau$ values in three datasets and (d-f) histograms of appearance pseudo-anomaly scores. In (a-c), brown vertical lines indicate the ratio of ground truth abnormal objects ($r$ in Table~\ref{table:datasets}). Black dashed and dotted horizontal lines are the performances of SOTA methods in OCVAD~\cite{jigsaw2022eccv} and UVAD~\cite{spr2022cvpr}, respectively. }
    \label{fig:varying_tau}
\end{figure}

We examined the impact of object cleansing on anomaly detection performance.
In Fig.~\ref{fig:varying_tau}(a-c), we illustrate the performances for Ped2, AVE, and STC datasets in partial mode using various $\tau$ values, where a $\tau$ value of 0 indicates no cleansing, and the larger $\tau$ value implies the stronger cleansing was applied.
Our results show that the detection performance improves as $\tau$ increases and starts to saturate at larger values.
From the steep increase in performance near $\tau=0$, we can conclude that cleansing plays a significant role in the proposed method.

\subsection{Detailed analysis of CKNN} \label{sec:analysis}

\subsubsection{Robustness to $\tau$ value}

Fig.~\ref{fig:varying_tau}(a-c) showed the detection performance approaches and surpasses the performance of UVAD SOTA method~\cite{spr2022cvpr} (black \textit{dotted} line) as $\tau$ approaches $r$.
Notably, we found that similar $\tau$ values of around 20 for the three datasets with different $r$ values (10.3\%, 4.1\%, and 12.6\%) yield accurate detection performance comparable to those of the OCVAD SOTA method~\cite{jigsaw2022eccv}, as depicted by the black \textit{dashed} lines.
Since $r$ is unknown at training time and the performance saturates for large $\tau$ values, choosing reasonably large $\tau$ value, around 20, is expected to yield satisfactory results.
To provide a systematic guideline for selecting the value of $\tau$, we suggest the following approach.

\begin{table}[t]
\setlength\tabcolsep{2pt}
  \centering
  \begin{minipage}{0.48\linewidth}
  \centering
\caption{OCVAD methods \\ in UVAD scenario.}
\vspace{-5pt}
\centering
  \begin{tabular}{c | c c c c}
   \hline
           & & Ped2 & AVE & STC \\
    \hline    
    \hline
    
\multirow{3}{*}{\cite{ssmt2021cvpr}}  & OC &  99.8 & 92.8 & 90.2 \\
\cdashline{2-5}
& P &  94.6 & 72.6 & 77.5 \\
& M & 96.0 & 75.0 & 79.3 \\
\hline
\multirow{3}{*}{\cite{jigsaw2022eccv}}  & OC & 99.9 & 93.0 & 90.6 \\
\cdashline{2-5}
  & P & 61.6 & 79.3 & 64.8 \\
& M & 81.6 & 86.4 & 78.2 \\
\hline
\multirow{2}{*}{\textbf{Ours}}   & P & \textbf{99.8} &  \textbf{94.0} &  \textbf{88.7}   \\    
                             & M & \textbf{99.7} & \textbf{94.1} & \textbf{89.0} \\  
\hline
\end{tabular}%
\centering
     \centering
  \label{table:ocvad}
  \end{minipage}   
  \begin{minipage}{0.48\linewidth}
\centering
\setlength\tabcolsep{2pt}    
{\renewcommand{\arraystretch}{1.25}%
\small
\centering
\captionof{table}{Ablation study using various components.}
\vspace{-5pt}
\label{table:ablation}
\centering
\begin{tabular}{c|c c} 
\hline %
 & P & M \\ %
 \hline \hline %
SOTA~\cite{spr2022cvpr} & 83.0 & 82.0 \\  \hline %
$\phi^{app}=$ ResNet & 88.6 & 88.9 \\ \hline
$f^{app}=$ kNN & 87.9 & 88.0 \\
$f^{app}=$ GMM & 87.2 & 88.3 \\ \hline
$f^{mot}=$ kNN & 88.7 & 87.8 \\
$f^{mot}=$ AE & 84.0 & 85.0 \\ \hline
\textbf{CKNN (Ours)} & \textbf{88.7} & \textbf{89.0} \\  \hline
\end{tabular}
}
\end{minipage}
\end{table}

\subsubsection{Selecting appropriate $\tau$ value}
In Fig.~\ref{fig:varying_tau}(d-f), we present histograms of appearance pseudo-anomaly scores for training objects in three datasets, and the scores exhibit a long-tailed distribution.
The dashed vertical lines in the plots indicate the pseudo-anomaly score values of corresponding percentiles, and the red dashed lines marked with $\tau^*$ denote where the tail of histogram begins.
Fig.~\ref{fig:varying_tau}(a-c) demonstrates that employing $\tau$ values larger than $\tau^*$ produces sufficiently high detection performance.
For instance, in Fig.~\ref{fig:varying_tau}(a), the AUROC for Ped2 dataset  exhibits near-perfect performance for $\tau$ larger than $\tau^*=10$ presented in Fig.~\ref{fig:varying_tau}(d).
In a real-world scenario, we recommend selecting $\tau$ values greater than the percentile where the tail of the pseudo-anomaly score histogram starts.

\subsubsection{OCVAD methods in UVAD scenario} \label{sec:ocvad_to_uvad}
In the UVAD scenario, the inclusion of anomalies within the training data mandates the development of methodologies that explicitly address this complication.
Table~\ref{table:ocvad} shows the performance of the two most SOTA OCVAD methods among those publicly available~\cite{ssmt2021cvpr,jigsaw2022eccv} when evaluated directly in the UVAD scenarios.
OC, P, and M refer to OCVAD, UVAD partial mode, and UVAD merge mode, respectively.
The results indicate significant performance degradation due to the anomalies present in the training data, highlighting the necessity for a UVAD-specific VAD method such as CKNN.

\subsubsection{Ablation Study 1: Employing different modules} \label{sec:ablation1}
Object cleansing, the core process of CKNN, relies on the choice of feature extractors and pseudo-anomaly score estimators.
We present CKNN performance evaluated on STC with altering components in Table~\ref{table:ablation}.
P and M indicate the partial and merge mode evaluation, respectively.
The proposed method retains consistent performance when substituting the appearance feature extractor from $\phi^{app}=$ CLIP to $\phi^{app}=$ ResNet~\cite{resnet}.
When examining the appearance pseudo-anomaly score estimators ($f^{app}$), using k-NN algorithm of $k=4$ using CLIP features or GMM with $n=8$ yield analogous results, both surpassing the SOTA method~\cite{spr2022cvpr}.
Furthermore, our trials with two alternative motion pseudo-anomaly score estimator ($f^{mot}$)---one utilizing k-NN based on optical flow features~\cite{attr2022arxiv} and the other leveraging AE reconstruction loss of optical flow map for individual objects---reinforced this consistent performance.
These results underscore the inherent robustness of object cleansing approach, making it less sensitive to specific component choices.

\begin{figure}[t]
\centering
\begin{minipage}{0.47\linewidth}
\centering
    \includegraphics[width=0.9\linewidth]{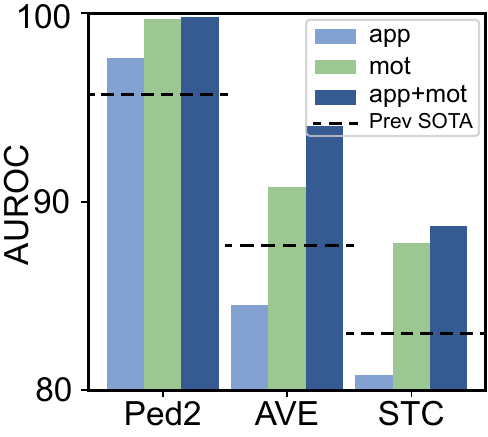}
    \vspace{-10pt}
    \caption{The contribution of appearance and motion-based anomaly scores.}
    \label{fig:ablation}
\end{minipage}
\begin{minipage}{0.45\linewidth}
  \centering
  \

\setlength\tabcolsep{2pt}    
{\renewcommand{\arraystretch}{1.2}%
\captionof{table}{Employing coreset \\ selection method~\cite{agarwal2005geometric}.} \label{table:coreset}
\vspace{-10pt}
\begin{tabular}{c | c c c}   
   \hline
           & & AVE & STC \\
    \hline    
    \hline
    
Random  & P & 94.0 & 88.7 \\
($p=1$) & M & 94.1 & 89.0 \\
\hline
Coreset~\cite{agarwal2005geometric} & P & 93.6 &  88.5   \\    
 ($p=1$)    & M & 93.9 & 88.8 \\  
\hline
No  & P &  93.8 & 88.8 \\
($p=100$) & M & 94.3 & 88.8 \\
\hline
\end{tabular}%
}
\centering
\end{minipage}       
\end{figure}

\subsubsection{Ablation Study 2: Appearance and motion} \label{sec:app_and_mot}
CKNN processes and evaluates objects based on two criteria: appearance and motion.
We present the detection performance of CKNN using individual scores in partial mode in Fig.~\ref{fig:ablation}.
The black dashed lines in each dataset indicate the performance of the UVAD SOTA method~\cite{spr2022cvpr}.
Remarkably, using a single score alone provides results that rival the SOTA method.
When both scores are combined, there is a notable enhancement in performance.
The synergistic effect of appearance and motion cleansing is in alignment with the observation in Sec~\ref{sec:tsne}.
A similar pattern is evident in the results using merge mode, which can be found in the supplementary materials.

\subsubsection{Ablation Study 3: Feature bank compression} \label{sec:compression}
For the feature bank compression illustrated in Eq.~\ref{eq:memorybank}, we conducted additional experiments using greedy coreset selection~\cite{agarwal2005geometric}.
Table~\ref{table:coreset} presents the performance of CKNN when feature bank compression is performed using random sampling with $p=1$ and coreset selection~\cite{agarwal2005geometric}.
Given the negligible performance difference from random subsampling, we opted to use random subsampling.

\begin{table}[t]
  \centering
     \setlength\tabcolsep{5pt}
  \caption{Throughput (FPS) of NN search in CKNN. FPS values are measured using the STC dataset in merge mode.}
  \vspace{-15pt}
  \label{table:runtime}
   \begin{tabular}{l c c}
   \\
   \hline
   \\[-1.3em]
         & $p=1$ & $p=100$ \\     
    \hline    
    \hline
    \\[-1em]        
    appearance & 115 & 115  \\
    motion & 8160 &  361  \\    
    
    \hline
    \end{tabular}%
    \vspace{-5pt}
     \centering
\end{table}

\subsubsection{Running time} \label{sec:runningtime}
For the successful integration of a VAD method into a real-time application, throughput is a critical constraint.
During the inference phase, CKNN executes three key steps: object detection, feature extraction, and NN search. 
Object detection and feature extraction using pretrained networks are widely adopted in other VAD methods~\cite{ssmt2021cvpr,jigsaw2022eccv,pretexts2023iccv,objae2019cvpr,future2018cvpr,appmot2019iccv,appmot2021aaai,spr2022cvpr}, but NN search can become a throughput bottleneck depending on the size of the training data.
We report the throughput of the NN search in frames per second (FPS) in Table~\ref{table:runtime}.
The measurement of FPS was conducted using STC dataset in merge mode, which is the most computationally heavy scenario; it has the largest number of NN search targets (more than 1 million).
Given that typical videos have an FPS of around 25, the NN search incurs negligible computational overhead even without feature bank compression.
Additionally, since feature bank compression has minimal impact on performance (as detailed in Sec~\ref{sec:robustness}), our method can achieve speeds of over 100 FPS while maintaining high performance, even on datasets that are 100 times larger.
Therefore, deploying CKNN in real-time is a viable solution.
For the experiments, the k-NN was implemented using a GPU~\cite{faiss} on a machine with a single NVIDIA GeForce RTX 2080 Ti GPU and an Intel i9-10940X @3.30GHz CPU.

\subsubsection{Robustness to $k$, $n$, and $p$ values} \label{sec:robustness}
\begin{figure}
    \centering
    \includegraphics[width=\linewidth]{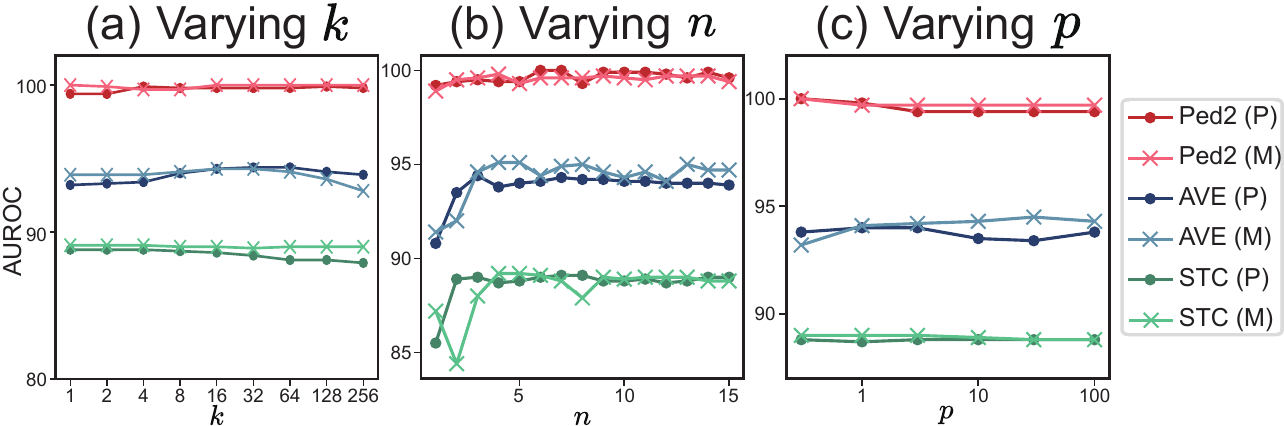}    
    \vspace{-15pt}
    \caption{Anomaly detection performances for varying values of (a) $k$ (b) $n$, and (c) $p$. Note that P and M indicate partial and merge mode, respectively.}
    \label{fig:varying_n_and_p}
    \vspace{-5pt}
\end{figure}

In Fig.~\ref{fig:varying_n_and_p}, we present the anomaly detection performances of CKNN for various values of $k$, $n$, and $p$.
Notably, the influence of these parameters on detection performance is limited; the performance remains consistent for $n$ values greater than three and across all $k$ values.
In addition, Fig.~\ref{fig:varying_n_and_p}(c) indicates that the $p$ exerts minimal influence on the overall detection performance.
Even as $p$ decreases from 100 (indicating no compression) to 0.3, the anomaly detection performances across all datasets remain consistent.
It's noteworthy that previous studies on IAD~\cite{patchcore2022cvpr,soft2022neurips} employed a $p$ value of 1.
Hence, we can infer that CKNN remains resilient irrespective of the $p$ value chosen, and using $p = 1$ leads to negligible performance deterioration.

\section{Conclusion}

In this paper, we addressed a problem arising in unsupervised video anomaly detection.
We tackled the problem using a k-NN algorithm which compares query features from a feature bank of training data.
One of the biggest challenges in this task is the presence of anomaly clusters in the training data, as it can mislead the anomaly detection method to confuse an anomaly for normal.
We overcame this issue by applying cleansing scheme prior to k-NN search.
The object cleansing takes place in two aspects, \ie, appearance and motion, and the benefits of this approach are supported by various experimental results and analyses.

Using those techniques, we proposed a novel method named CKNN and achieved the state-of-the-art performances on various UVAD datasets.
The detection performances are even comparable to those of the state-of-the-art method in OCVAD~\cite{jigsaw2022eccv}, which learns with anomaly-free data.
This implies that our method can perform as well as OCVAD method with noisy data.
We believe that, leveraging our method, system managers can now build an automated VAD system with much less effort on data preparation.

\begin{acks}
This work was supported by the BK21 FOUR program of the Education and Research Program for Future ICT Pioneers, Seoul National University in 2024, Institute of Information \& Communications Technology Planning \& Evaluation (IITP) grant funded by the Korea government (MSIT) [NO.2021-0-01343, Artificial Intelligence Graduate School Program (Seoul National University)], the National Research Foundation of Korea (NRF) grant funded by the Korea government (MSIT) (No. 2022R1A3B1077720), the National Research Foundation of Korea (NRF) grant funded by the Korea government (MSIT) (No. 2022R1A5A708390811), and Samsung Electronics.
\end{acks}




\bibliographystyle{ACM-Reference-Format}
\bibliography{aaai24}


\begin{thebibliography}{51}


\ifx \showCODEN    \undefined \def \showCODEN     #1{\unskip}     \fi
\ifx \showDOI      \undefined \def \showDOI       #1{#1}\fi
\ifx \showISBNx    \undefined \def \showISBNx     #1{\unskip}     \fi
\ifx \showISBNxiii \undefined \def \showISBNxiii  #1{\unskip}     \fi
\ifx \showISSN     \undefined \def \showISSN      #1{\unskip}     \fi
\ifx \showLCCN     \undefined \def \showLCCN      #1{\unskip}     \fi
\ifx \shownote     \undefined \def \shownote      #1{#1}          \fi
\ifx \showarticletitle \undefined \def \showarticletitle #1{#1}   \fi
\ifx \showURL      \undefined \def \showURL       {\relax}        \fi
\providecommand\bibfield[2]{#2}
\providecommand\bibinfo[2]{#2}
\providecommand\natexlab[1]{#1}
\providecommand\showeprint[2][]{arXiv:#2}

\bibitem[Acsintoae et~al\mbox{.}(2022)]%
        {ubnormal}
\bibfield{author}{\bibinfo{person}{Andra Acsintoae}, \bibinfo{person}{Andrei Florescu}, \bibinfo{person}{Mariana-Iuliana Georgescu}, \bibinfo{person}{Tudor Mare}, \bibinfo{person}{Paul Sumedrea}, \bibinfo{person}{Radu~Tudor Ionescu}, \bibinfo{person}{Fahad~Shahbaz Khan}, {and} \bibinfo{person}{Mubarak Shah}.} \bibinfo{year}{2022}\natexlab{}.
\newblock \showarticletitle{Ubnormal: New benchmark for supervised open-set video anomaly detection}. In \bibinfo{booktitle}{\emph{Proceedings of the IEEE/CVF conference on computer vision and pattern recognition}}. \bibinfo{pages}{20143--20153}.
\newblock


\bibitem[Agarwal et~al\mbox{.}(2005)]%
        {agarwal2005geometric}
\bibfield{author}{\bibinfo{person}{Pankaj~K Agarwal}, \bibinfo{person}{Sariel Har-Peled}, \bibinfo{person}{Kasturi~R Varadarajan}, {et~al\mbox{.}}} \bibinfo{year}{2005}\natexlab{}.
\newblock \showarticletitle{Geometric approximation via coresets}.
\newblock \bibinfo{journal}{\emph{Combinatorial and computational geometry}} \bibinfo{volume}{52}, \bibinfo{number}{1} (\bibinfo{year}{2005}), \bibinfo{pages}{1--30}.
\newblock


\bibitem[Akcay et~al\mbox{.}(2019)]%
        {akcay2019ganomaly}
\bibfield{author}{\bibinfo{person}{Samet Akcay}, \bibinfo{person}{Amir Atapour-Abarghouei}, {and} \bibinfo{person}{Toby~P Breckon}.} \bibinfo{year}{2019}\natexlab{}.
\newblock \showarticletitle{Ganomaly: Semi-supervised anomaly detection via adversarial training}. In \bibinfo{booktitle}{\emph{Computer Vision--ACCV 2018: 14th Asian Conference on Computer Vision, Perth, Australia, December 2--6, 2018, Revised Selected Papers, Part III 14}}. Springer, \bibinfo{pages}{622--637}.
\newblock


\bibitem[Bergman et~al\mbox{.}(2020)]%
        {dn22020}
\bibfield{author}{\bibinfo{person}{Liron Bergman}, \bibinfo{person}{Niv Cohen}, {and} \bibinfo{person}{Yedid Hoshen}.} \bibinfo{year}{2020}\natexlab{}.
\newblock \showarticletitle{Deep nearest neighbor anomaly detection}.
\newblock \bibinfo{journal}{\emph{arXiv preprint arXiv:2002.10445}} (\bibinfo{year}{2020}).
\newblock


\bibitem[Cai et~al\mbox{.}(2021)]%
        {appmot2021aaai}
\bibfield{author}{\bibinfo{person}{Ruichu Cai}, \bibinfo{person}{Hao Zhang}, \bibinfo{person}{Wen Liu}, \bibinfo{person}{Shenghua Gao}, {and} \bibinfo{person}{Zhifeng Hao}.} \bibinfo{year}{2021}\natexlab{}.
\newblock \showarticletitle{Appearance-motion memory consistency network for video anomaly detection}. In \bibinfo{booktitle}{\emph{AAAI}}, Vol.~\bibinfo{volume}{35}. \bibinfo{pages}{938--946}.
\newblock


\bibitem[Chan and Vasconcelos(2008)]%
        {ped2}
\bibfield{author}{\bibinfo{person}{Antoni~B Chan} {and} \bibinfo{person}{Nuno Vasconcelos}.} \bibinfo{year}{2008}\natexlab{}.
\newblock \showarticletitle{Modeling, clustering, and segmenting video with mixtures of dynamic textures}.
\newblock \bibinfo{journal}{\emph{IEEE transactions on pattern analysis and machine intelligence}} \bibinfo{volume}{30}, \bibinfo{number}{5} (\bibinfo{year}{2008}), \bibinfo{pages}{909--926}.
\newblock


\bibitem[Cohen and Hoshen(2020)]%
        {spade2020}
\bibfield{author}{\bibinfo{person}{Niv Cohen} {and} \bibinfo{person}{Yedid Hoshen}.} \bibinfo{year}{2020}\natexlab{}.
\newblock \showarticletitle{Sub-image anomaly detection with deep pyramid correspondences}.
\newblock \bibinfo{journal}{\emph{arXiv preprint arXiv:2005.02357}} (\bibinfo{year}{2020}).
\newblock


\bibitem[Del~Giorno et~al\mbox{.}(2016)]%
        {discriminative2016eccv}
\bibfield{author}{\bibinfo{person}{Allison Del~Giorno}, \bibinfo{person}{J~Andrew Bagnell}, {and} \bibinfo{person}{Martial Hebert}.} \bibinfo{year}{2016}\natexlab{}.
\newblock \showarticletitle{A discriminative framework for anomaly detection in large videos}. In \bibinfo{booktitle}{\emph{ECCV}}. Springer, \bibinfo{pages}{334--349}.
\newblock


\bibitem[Dosovitskiy et~al\mbox{.}(2015)]%
        {flownet}
\bibfield{author}{\bibinfo{person}{Alexey Dosovitskiy}, \bibinfo{person}{Philipp Fischer}, \bibinfo{person}{Eddy Ilg}, \bibinfo{person}{Philip Hausser}, \bibinfo{person}{Caner Hazirbas}, \bibinfo{person}{Vladimir Golkov}, \bibinfo{person}{Patrick Van Der~Smagt}, \bibinfo{person}{Daniel Cremers}, {and} \bibinfo{person}{Thomas Brox}.} \bibinfo{year}{2015}\natexlab{}.
\newblock \showarticletitle{Flownet: Learning optical flow with convolutional networks}. In \bibinfo{booktitle}{\emph{ICCV}}. \bibinfo{pages}{2758--2766}.
\newblock


\bibitem[Eskin et~al\mbox{.}(2002)]%
        {knn}
\bibfield{author}{\bibinfo{person}{Eleazar Eskin}, \bibinfo{person}{Andrew Arnold}, \bibinfo{person}{Michael Prerau}, \bibinfo{person}{Leonid Portnoy}, {and} \bibinfo{person}{Sal Stolfo}.} \bibinfo{year}{2002}\natexlab{}.
\newblock \showarticletitle{A geometric framework for unsupervised anomaly detection: Detecting intrusions in unlabeled data}.
\newblock \bibinfo{journal}{\emph{Applications of data mining in computer security}} (\bibinfo{year}{2002}), \bibinfo{pages}{77--101}.
\newblock


\bibitem[Georgescu et~al\mbox{.}(2021)]%
        {ssmt2021cvpr}
\bibfield{author}{\bibinfo{person}{Mariana-Iuliana Georgescu}, \bibinfo{person}{Antonio Barbalau}, \bibinfo{person}{Radu~Tudor Ionescu}, \bibinfo{person}{Fahad~Shahbaz Khan}, \bibinfo{person}{Marius Popescu}, {and} \bibinfo{person}{Mubarak Shah}.} \bibinfo{year}{2021}\natexlab{}.
\newblock \showarticletitle{Anomaly detection in video via self-supervised and multi-task learning}. In \bibinfo{booktitle}{\emph{CVPR}}. \bibinfo{pages}{12742--12752}.
\newblock


\bibitem[Gong et~al\mbox{.}(2019)]%
        {memae2019iccv}
\bibfield{author}{\bibinfo{person}{Dong Gong}, \bibinfo{person}{Lingqiao Liu}, \bibinfo{person}{Vuong Le}, \bibinfo{person}{Budhaditya Saha}, \bibinfo{person}{Moussa~Reda Mansour}, \bibinfo{person}{Svetha Venkatesh}, {and} \bibinfo{person}{Anton van~den Hengel}.} \bibinfo{year}{2019}\natexlab{}.
\newblock \showarticletitle{Memorizing normality to detect anomaly: Memory-augmented deep autoencoder for unsupervised anomaly detection}. In \bibinfo{booktitle}{\emph{ICCV}}. \bibinfo{pages}{1705--1714}.
\newblock


\bibitem[Gu et~al\mbox{.}(2019)]%
        {knn_tabular}
\bibfield{author}{\bibinfo{person}{Xiaoyi Gu}, \bibinfo{person}{Leman Akoglu}, {and} \bibinfo{person}{Alessandro Rinaldo}.} \bibinfo{year}{2019}\natexlab{}.
\newblock \showarticletitle{Statistical analysis of nearest neighbor methods for anomaly detection}.
\newblock \bibinfo{journal}{\emph{NeurIPS}}  \bibinfo{volume}{32} (\bibinfo{year}{2019}).
\newblock


\bibitem[He et~al\mbox{.}(2017)]%
        {maskrcnn}
\bibfield{author}{\bibinfo{person}{Kaiming He}, \bibinfo{person}{Georgia Gkioxari}, \bibinfo{person}{Piotr Doll{\'a}r}, {and} \bibinfo{person}{Ross Girshick}.} \bibinfo{year}{2017}\natexlab{}.
\newblock \showarticletitle{Mask r-cnn}. In \bibinfo{booktitle}{\emph{ICCV}}. \bibinfo{pages}{2961--2969}.
\newblock


\bibitem[He et~al\mbox{.}(2016)]%
        {resnet}
\bibfield{author}{\bibinfo{person}{Kaiming He}, \bibinfo{person}{Xiangyu Zhang}, \bibinfo{person}{Shaoqing Ren}, {and} \bibinfo{person}{Jian Sun}.} \bibinfo{year}{2016}\natexlab{}.
\newblock \showarticletitle{Deep residual learning for image recognition}. In \bibinfo{booktitle}{\emph{CVPR}}. \bibinfo{pages}{770--778}.
\newblock


\bibitem[Hyun et~al\mbox{.}(2024)]%
        {hyun2024reconpatch}
\bibfield{author}{\bibinfo{person}{Jeeho Hyun}, \bibinfo{person}{Sangyun Kim}, \bibinfo{person}{Giyoung Jeon}, \bibinfo{person}{Seung~Hwan Kim}, \bibinfo{person}{Kyunghoon Bae}, {and} \bibinfo{person}{Byung~Jun Kang}.} \bibinfo{year}{2024}\natexlab{}.
\newblock \showarticletitle{ReConPatch: Contrastive patch representation learning for industrial anomaly detection}. In \bibinfo{booktitle}{\emph{Proceedings of the IEEE/CVF Winter Conference on Applications of Computer Vision}}. \bibinfo{pages}{2052--2061}.
\newblock


\bibitem[Ilg et~al\mbox{.}(2017)]%
        {flownet2}
\bibfield{author}{\bibinfo{person}{Eddy Ilg}, \bibinfo{person}{Nikolaus Mayer}, \bibinfo{person}{Tonmoy Saikia}, \bibinfo{person}{Margret Keuper}, \bibinfo{person}{Alexey Dosovitskiy}, {and} \bibinfo{person}{Thomas Brox}.} \bibinfo{year}{2017}\natexlab{}.
\newblock \showarticletitle{Flownet 2.0: Evolution of optical flow estimation with deep networks}. In \bibinfo{booktitle}{\emph{CVPR}}. \bibinfo{pages}{2462--2470}.
\newblock


\bibitem[Ionescu et~al\mbox{.}(2019a)]%
        {objae2019cvpr}
\bibfield{author}{\bibinfo{person}{Radu~Tudor Ionescu}, \bibinfo{person}{Fahad~Shahbaz Khan}, \bibinfo{person}{Mariana-Iuliana Georgescu}, {and} \bibinfo{person}{Ling Shao}.} \bibinfo{year}{2019}\natexlab{a}.
\newblock \showarticletitle{Object-centric auto-encoders and dummy anomalies for abnormal event detection in video}. In \bibinfo{booktitle}{\emph{CVPR}}. \bibinfo{pages}{7842--7851}.
\newblock


\bibitem[Ionescu et~al\mbox{.}(2019b)]%
        {narrowed}
\bibfield{author}{\bibinfo{person}{Radu~Tudor Ionescu}, \bibinfo{person}{Sorina Smeureanu}, \bibinfo{person}{Marius Popescu}, {and} \bibinfo{person}{Bogdan Alexe}.} \bibinfo{year}{2019}\natexlab{b}.
\newblock \showarticletitle{Detecting abnormal events in video using narrowed normality clusters}. In \bibinfo{booktitle}{\emph{2019 IEEE winter conference on applications of computer vision (WACV)}}. IEEE, \bibinfo{pages}{1951--1960}.
\newblock


\bibitem[Jeong et~al\mbox{.}(2023)]%
        {jeong2023winclip}
\bibfield{author}{\bibinfo{person}{Jongheon Jeong}, \bibinfo{person}{Yang Zou}, \bibinfo{person}{Taewan Kim}, \bibinfo{person}{Dongqing Zhang}, \bibinfo{person}{Avinash Ravichandran}, {and} \bibinfo{person}{Onkar Dabeer}.} \bibinfo{year}{2023}\natexlab{}.
\newblock \showarticletitle{Winclip: Zero-/few-shot anomaly classification and segmentation}. In \bibinfo{booktitle}{\emph{Proceedings of the IEEE/CVF Conference on Computer Vision and Pattern Recognition}}. \bibinfo{pages}{19606--19616}.
\newblock


\bibitem[Johnson et~al\mbox{.}(2019)]%
        {faiss}
\bibfield{author}{\bibinfo{person}{Jeff Johnson}, \bibinfo{person}{Matthijs Douze}, {and} \bibinfo{person}{Herv{\'e} J{\'e}gou}.} \bibinfo{year}{2019}\natexlab{}.
\newblock \showarticletitle{Billion-scale similarity search with {GPUs}}.
\newblock \bibinfo{journal}{\emph{IEEE Transactions on Big Data}} \bibinfo{volume}{7}, \bibinfo{number}{3} (\bibinfo{year}{2019}), \bibinfo{pages}{535--547}.
\newblock


\bibitem[Kingma and Ba(2014)]%
        {adam}
\bibfield{author}{\bibinfo{person}{Diederik~P Kingma} {and} \bibinfo{person}{Jimmy Ba}.} \bibinfo{year}{2014}\natexlab{}.
\newblock \showarticletitle{Adam: A method for stochastic optimization}.
\newblock \bibinfo{journal}{\emph{arXiv preprint arXiv:1412.6980}} (\bibinfo{year}{2014}).
\newblock


\bibitem[Liu et~al\mbox{.}(2024)]%
        {ucad}
\bibfield{author}{\bibinfo{person}{Jiaqi Liu}, \bibinfo{person}{Kai Wu}, \bibinfo{person}{Qiang Nie}, \bibinfo{person}{Ying Chen}, \bibinfo{person}{Bin-Bin Gao}, \bibinfo{person}{Yong Liu}, \bibinfo{person}{Jinbao Wang}, \bibinfo{person}{Chengjie Wang}, {and} \bibinfo{person}{Feng Zheng}.} \bibinfo{year}{2024}\natexlab{}.
\newblock \showarticletitle{Unsupervised Continual Anomaly Detection with Contrastively-learned Prompt}.
\newblock \bibinfo{journal}{\emph{arXiv preprint arXiv:2401.01010}} (\bibinfo{year}{2024}).
\newblock


\bibitem[Liu et~al\mbox{.}(2018b)]%
        {future2018cvpr}
\bibfield{author}{\bibinfo{person}{Wen Liu}, \bibinfo{person}{Weixin Luo}, \bibinfo{person}{Dongze Lian}, {and} \bibinfo{person}{Shenghua Gao}.} \bibinfo{year}{2018}\natexlab{b}.
\newblock \showarticletitle{Future frame prediction for anomaly detection--a new baseline}. In \bibinfo{booktitle}{\emph{CVPR}}. \bibinfo{pages}{6536--6545}.
\newblock


\bibitem[Liu et~al\mbox{.}(2018c)]%
        {stc}
\bibfield{author}{\bibinfo{person}{W. Liu}, \bibinfo{person}{D.~Lian W.~Luo}, {and} \bibinfo{person}{S. Gao}.} \bibinfo{year}{2018}\natexlab{c}.
\newblock \showarticletitle{Future Frame Prediction for Anomaly Detection -- A New Baseline}. In \bibinfo{booktitle}{\emph{2018 IEEE Conference on Computer Vision and Pattern Recognition (CVPR)}}.
\newblock


\bibitem[Liu et~al\mbox{.}(2018a)]%
        {classifier2018bmvc}
\bibfield{author}{\bibinfo{person}{Yusha Liu}, \bibinfo{person}{Chun-Liang Li}, {and} \bibinfo{person}{Barnab{\'a}s P{\'o}czos}.} \bibinfo{year}{2018}\natexlab{a}.
\newblock \showarticletitle{Classifier Two Sample Test for Video Anomaly Detections.}. In \bibinfo{booktitle}{\emph{BMVC}}. \bibinfo{pages}{71}.
\newblock


\bibitem[Liu et~al\mbox{.}(2021)]%
        {flowguided2021iccv}
\bibfield{author}{\bibinfo{person}{Zhian Liu}, \bibinfo{person}{Yongwei Nie}, \bibinfo{person}{Chengjiang Long}, \bibinfo{person}{Qing Zhang}, {and} \bibinfo{person}{Guiqing Li}.} \bibinfo{year}{2021}\natexlab{}.
\newblock \showarticletitle{A hybrid video anomaly detection framework via memory-augmented flow reconstruction and flow-guided frame prediction}. In \bibinfo{booktitle}{\emph{ICCV}}. \bibinfo{pages}{13588--13597}.
\newblock


\bibitem[Lu et~al\mbox{.}(2013)]%
        {avenue}
\bibfield{author}{\bibinfo{person}{Cewu Lu}, \bibinfo{person}{Jianping Shi}, {and} \bibinfo{person}{Jiaya Jia}.} \bibinfo{year}{2013}\natexlab{}.
\newblock \showarticletitle{Abnormal event detection at 150 fps in matlab}. In \bibinfo{booktitle}{\emph{Proceedings of the IEEE international conference on computer vision}}. \bibinfo{pages}{2720--2727}.
\newblock


\bibitem[Nguyen and Meunier(2019)]%
        {appmot2019iccv}
\bibfield{author}{\bibinfo{person}{Trong-Nguyen Nguyen} {and} \bibinfo{person}{Jean Meunier}.} \bibinfo{year}{2019}\natexlab{}.
\newblock \showarticletitle{Anomaly detection in video sequence with appearance-motion correspondence}. In \bibinfo{booktitle}{\emph{ICCV}}. \bibinfo{pages}{1273--1283}.
\newblock


\bibitem[Pang et~al\mbox{.}(2020)]%
        {ordinal2020cvpr}
\bibfield{author}{\bibinfo{person}{Guansong Pang}, \bibinfo{person}{Cheng Yan}, \bibinfo{person}{Chunhua Shen}, \bibinfo{person}{Anton van~den Hengel}, {and} \bibinfo{person}{Xiao Bai}.} \bibinfo{year}{2020}\natexlab{}.
\newblock \showarticletitle{Self-trained deep ordinal regression for end-to-end video anomaly detection}. In \bibinfo{booktitle}{\emph{CVPR}}. \bibinfo{pages}{12173--12182}.
\newblock


\bibitem[Park et~al\mbox{.}(2020)]%
        {mnad2020cvpr}
\bibfield{author}{\bibinfo{person}{Hyunjong Park}, \bibinfo{person}{Jongyoun Noh}, {and} \bibinfo{person}{Bumsub Ham}.} \bibinfo{year}{2020}\natexlab{}.
\newblock \showarticletitle{Learning memory-guided normality for anomaly detection}. In \bibinfo{booktitle}{\emph{CVPR}}. \bibinfo{pages}{14372--14381}.
\newblock


\bibitem[Paszke et~al\mbox{.}(2019)]%
        {pytorch}
\bibfield{author}{\bibinfo{person}{Adam Paszke}, \bibinfo{person}{Sam Gross}, \bibinfo{person}{Francisco Massa}, \bibinfo{person}{Adam Lerer}, \bibinfo{person}{James Bradbury}, \bibinfo{person}{Gregory Chanan}, \bibinfo{person}{Trevor Killeen}, \bibinfo{person}{Zeming Lin}, \bibinfo{person}{Natalia Gimelshein}, \bibinfo{person}{Luca Antiga}, \bibinfo{person}{Alban Desmaison}, \bibinfo{person}{Andreas Kopf}, \bibinfo{person}{Edward Yang}, \bibinfo{person}{Zachary DeVito}, \bibinfo{person}{Martin Raison}, \bibinfo{person}{Alykhan Tejani}, \bibinfo{person}{Sasank Chilamkurthy}, \bibinfo{person}{Benoit Steiner}, \bibinfo{person}{Lu Fang}, \bibinfo{person}{Junjie Bai}, {and} \bibinfo{person}{Soumith Chintala}.} \bibinfo{year}{2019}\natexlab{}.
\newblock \showarticletitle{{PyTorch: An Imperative Style, High-Performance Deep Learning Library}}. In \bibinfo{booktitle}{\emph{Advances in Neural Information Processing Systems 32}}, \bibfield{editor}{\bibinfo{person}{H.~Wallach}, \bibinfo{person}{H.~Larochelle}, \bibinfo{person}{A.~Beygelzimer}, \bibinfo{person}{F.~d'Alché Buc}, \bibinfo{person}{E.~Fox}, {and} \bibinfo{person}{R.~Garnett}} (Eds.). \bibinfo{publisher}{Curran Associates, Inc.}, \bibinfo{pages}{8024--8035}.
\newblock
\urldef\tempurl%
\url{http://papers.neurips.cc/paper/9015-pytorch-an-imperative-style-high-performance-deep-learning-library.pdf}
\showURL{%
\tempurl}


\bibitem[Pedregosa et~al\mbox{.}(2011)]%
        {scikitlearn}
\bibfield{author}{\bibinfo{person}{F. Pedregosa}, \bibinfo{person}{G. Varoquaux}, \bibinfo{person}{A. Gramfort}, \bibinfo{person}{V. Michel}, \bibinfo{person}{B. Thirion}, \bibinfo{person}{O. Grisel}, \bibinfo{person}{M. Blondel}, \bibinfo{person}{P. Prettenhofer}, \bibinfo{person}{R. Weiss}, \bibinfo{person}{V. Dubourg}, \bibinfo{person}{J. Vanderplas}, \bibinfo{person}{A. Passos}, \bibinfo{person}{D. Cournapeau}, \bibinfo{person}{M. Brucher}, \bibinfo{person}{M. Perrot}, {and} \bibinfo{person}{E. Duchesnay}.} \bibinfo{year}{2011}\natexlab{}.
\newblock \showarticletitle{Scikit-learn: Machine Learning in {P}ython}.
\newblock \bibinfo{journal}{\emph{Journal of Machine Learning Research}}  \bibinfo{volume}{12} (\bibinfo{year}{2011}), \bibinfo{pages}{2825--2830}.
\newblock


\bibitem[Perera et~al\mbox{.}(2019)]%
        {ocgan2019cvpr}
\bibfield{author}{\bibinfo{person}{Pramuditha Perera}, \bibinfo{person}{Ramesh Nallapati}, {and} \bibinfo{person}{Bing Xiang}.} \bibinfo{year}{2019}\natexlab{}.
\newblock \showarticletitle{Ocgan: One-class novelty detection using gans with constrained latent representations}. In \bibinfo{booktitle}{\emph{CVPR}}. \bibinfo{pages}{2898--2906}.
\newblock


\bibitem[Radford et~al\mbox{.}(2021)]%
        {clip}
\bibfield{author}{\bibinfo{person}{Alec Radford}, \bibinfo{person}{Jong~Wook Kim}, \bibinfo{person}{Chris Hallacy}, \bibinfo{person}{Aditya Ramesh}, \bibinfo{person}{Gabriel Goh}, \bibinfo{person}{Sandhini Agarwal}, \bibinfo{person}{Girish Sastry}, \bibinfo{person}{Amanda Askell}, \bibinfo{person}{Pamela Mishkin}, \bibinfo{person}{Jack Clark}, {et~al\mbox{.}}} \bibinfo{year}{2021}\natexlab{}.
\newblock \showarticletitle{Learning transferable visual models from natural language supervision}. In \bibinfo{booktitle}{\emph{ICML}}. PMLR, \bibinfo{pages}{8748--8763}.
\newblock


\bibitem[Redmon and Farhadi(2018)]%
        {yolov3}
\bibfield{author}{\bibinfo{person}{Joseph Redmon} {and} \bibinfo{person}{Ali Farhadi}.} \bibinfo{year}{2018}\natexlab{}.
\newblock \showarticletitle{Yolov3: An incremental improvement}.
\newblock \bibinfo{journal}{\emph{arXiv preprint arXiv:1804.02767}} (\bibinfo{year}{2018}).
\newblock


\bibitem[Reiss and Hoshen(2022)]%
        {attr2022arxiv}
\bibfield{author}{\bibinfo{person}{Tal Reiss} {and} \bibinfo{person}{Yedid Hoshen}.} \bibinfo{year}{2022}\natexlab{}.
\newblock \showarticletitle{Attribute-based Representations for Accurate and Interpretable Video Anomaly Detection}.
\newblock \bibinfo{journal}{\emph{arXiv preprint arXiv:2212.00789}} (\bibinfo{year}{2022}).
\newblock


\bibitem[Roth et~al\mbox{.}(2022)]%
        {patchcore2022cvpr}
\bibfield{author}{\bibinfo{person}{Karsten Roth}, \bibinfo{person}{Latha Pemula}, \bibinfo{person}{Joaquin Zepeda}, \bibinfo{person}{Bernhard Sch{\"o}lkopf}, \bibinfo{person}{Thomas Brox}, {and} \bibinfo{person}{Peter Gehler}.} \bibinfo{year}{2022}\natexlab{}.
\newblock \showarticletitle{Towards total recall in industrial anomaly detection}. In \bibinfo{booktitle}{\emph{CVPR}}. \bibinfo{pages}{14318--14328}.
\newblock


\bibitem[Ruff et~al\mbox{.}(2020)]%
        {deepsad}
\bibfield{author}{\bibinfo{person}{Lukas Ruff}, \bibinfo{person}{Robert~A Vandermeulen}, \bibinfo{person}{Nico G{\"o}rnitz}, \bibinfo{person}{Alexander Binder}, \bibinfo{person}{Emmanuel M{\"u}ller}, \bibinfo{person}{Klaus-Robert M{\"u}ller}, {and} \bibinfo{person}{Marius Kloft}.} \bibinfo{year}{2020}\natexlab{}.
\newblock \showarticletitle{Deep semi-supervised anomaly detection}. In \bibinfo{booktitle}{\emph{ICLR}}.
\newblock


\bibitem[Sabokrou et~al\mbox{.}(2018)]%
        {RnD}
\bibfield{author}{\bibinfo{person}{Mohammad Sabokrou}, \bibinfo{person}{Mohammad Khalooei}, \bibinfo{person}{Mahmood Fathy}, {and} \bibinfo{person}{Ehsan Adeli}.} \bibinfo{year}{2018}\natexlab{}.
\newblock \showarticletitle{Adversarially learned one-class classifier for novelty detection}. In \bibinfo{booktitle}{\emph{Proceedings of the IEEE conference on computer vision and pattern recognition}}. \bibinfo{pages}{3379--3388}.
\newblock


\bibitem[Shi et~al\mbox{.}(2023)]%
        {pretexts2023iccv}
\bibfield{author}{\bibinfo{person}{Chenrui Shi}, \bibinfo{person}{Che Sun}, \bibinfo{person}{Yuwei Wu}, {and} \bibinfo{person}{Yunde Jia}.} \bibinfo{year}{2023}\natexlab{}.
\newblock \showarticletitle{Video anomaly detection via sequentially learning multiple pretext tasks}. In \bibinfo{booktitle}{\emph{Proceedings of the IEEE/CVF International Conference on Computer Vision}}. \bibinfo{pages}{10330--10340}.
\newblock


\bibitem[Singh et~al\mbox{.}(2023)]%
        {eval2023cvpr}
\bibfield{author}{\bibinfo{person}{Ashish Singh}, \bibinfo{person}{Michael~J Jones}, {and} \bibinfo{person}{Erik~G Learned-Miller}.} \bibinfo{year}{2023}\natexlab{}.
\newblock \showarticletitle{EVAL: Explainable Video Anomaly Localization}. In \bibinfo{booktitle}{\emph{Proceedings of the IEEE/CVF Conference on Computer Vision and Pattern Recognition}}. \bibinfo{pages}{18717--18726}.
\newblock


\bibitem[Tudor~Ionescu et~al\mbox{.}(2017)]%
        {unmasking2017iccv}
\bibfield{author}{\bibinfo{person}{Radu Tudor~Ionescu}, \bibinfo{person}{Sorina Smeureanu}, \bibinfo{person}{Bogdan Alexe}, {and} \bibinfo{person}{Marius Popescu}.} \bibinfo{year}{2017}\natexlab{}.
\newblock \showarticletitle{Unmasking the abnormal events in video}. In \bibinfo{booktitle}{\emph{ICCV}}. \bibinfo{pages}{2895--2903}.
\newblock


\bibitem[Van~der Maaten and Hinton(2008)]%
        {tsne}
\bibfield{author}{\bibinfo{person}{Laurens Van~der Maaten} {and} \bibinfo{person}{Geoffrey Hinton}.} \bibinfo{year}{2008}\natexlab{}.
\newblock \showarticletitle{Visualizing data using t-SNE.}
\newblock \bibinfo{journal}{\emph{Journal of machine learning research}} \bibinfo{volume}{9}, \bibinfo{number}{11} (\bibinfo{year}{2008}).
\newblock


\bibitem[Wang et~al\mbox{.}(2022)]%
        {jigsaw2022eccv}
\bibfield{author}{\bibinfo{person}{Guodong Wang}, \bibinfo{person}{Yunhong Wang}, \bibinfo{person}{Jie Qin}, \bibinfo{person}{Dongming Zhang}, \bibinfo{person}{Xiuguo Bao}, {and} \bibinfo{person}{Di Huang}.} \bibinfo{year}{2022}\natexlab{}.
\newblock \showarticletitle{Video anomaly detection by solving decoupled spatio-temporal jigsaw puzzles}. In \bibinfo{booktitle}{\emph{ECCV}}. Springer, \bibinfo{pages}{494--511}.
\newblock


\bibitem[Xi et~al\mbox{.}(2022)]%
        {soft2022neurips}
\bibfield{author}{\bibinfo{person}{Jiang Xi}, \bibinfo{person}{Jianlin Liu}, \bibinfo{person}{Jinbao Wang}, \bibinfo{person}{Qiang Nie}, \bibinfo{person}{WU Kai}, \bibinfo{person}{Yong Liu}, \bibinfo{person}{Chengjie Wang}, {and} \bibinfo{person}{Feng Zheng}.} \bibinfo{year}{2022}\natexlab{}.
\newblock \showarticletitle{SoftPatch: Unsupervised Anomaly Detection with Noisy Data}. In \bibinfo{booktitle}{\emph{NeurIPS}}.
\newblock


\bibitem[Xu et~al\mbox{.}(2015)]%
        {leakyrelu}
\bibfield{author}{\bibinfo{person}{Bing Xu}, \bibinfo{person}{Naiyan Wang}, \bibinfo{person}{Tianqi Chen}, {and} \bibinfo{person}{Mu Li}.} \bibinfo{year}{2015}\natexlab{}.
\newblock \showarticletitle{Empirical evaluation of rectified activations in convolutional network}.
\newblock \bibinfo{journal}{\emph{arXiv preprint arXiv:1505.00853}} (\bibinfo{year}{2015}).
\newblock


\bibitem[Yan et~al\mbox{.}(2023)]%
        {diffusion2023iccv}
\bibfield{author}{\bibinfo{person}{Cheng Yan}, \bibinfo{person}{Shiyu Zhang}, \bibinfo{person}{Yang Liu}, \bibinfo{person}{Guansong Pang}, {and} \bibinfo{person}{Wenjun Wang}.} \bibinfo{year}{2023}\natexlab{}.
\newblock \showarticletitle{Feature prediction diffusion model for video anomaly detection}. In \bibinfo{booktitle}{\emph{Proceedings of the IEEE/CVF International Conference on Computer Vision}}. \bibinfo{pages}{5527--5537}.
\newblock


\bibitem[Yang et~al\mbox{.}(2023)]%
        {event2023cvpr}
\bibfield{author}{\bibinfo{person}{Zhiwei Yang}, \bibinfo{person}{Jing Liu}, \bibinfo{person}{Zhaoyang Wu}, \bibinfo{person}{Peng Wu}, {and} \bibinfo{person}{Xiaotao Liu}.} \bibinfo{year}{2023}\natexlab{}.
\newblock \showarticletitle{Video Event Restoration Based on Keyframes for Video Anomaly Detection}. In \bibinfo{booktitle}{\emph{Proceedings of the IEEE/CVF Conference on Computer Vision and Pattern Recognition}}. \bibinfo{pages}{14592--14601}.
\newblock


\bibitem[Yi and Yoon(2020)]%
        {patch2020accv}
\bibfield{author}{\bibinfo{person}{Jihun Yi} {and} \bibinfo{person}{Sungroh Yoon}.} \bibinfo{year}{2020}\natexlab{}.
\newblock \showarticletitle{Patch svdd: Patch-level svdd for anomaly detection and segmentation}. In \bibinfo{booktitle}{\emph{ACCV}}.
\newblock


\bibitem[Yu et~al\mbox{.}(2022)]%
        {spr2022cvpr}
\bibfield{author}{\bibinfo{person}{Guang Yu}, \bibinfo{person}{Siqi Wang}, \bibinfo{person}{Zhiping Cai}, \bibinfo{person}{Xinwang Liu}, \bibinfo{person}{Chuanfu Xu}, {and} \bibinfo{person}{Chengkun Wu}.} \bibinfo{year}{2022}\natexlab{}.
\newblock \showarticletitle{Deep anomaly discovery from unlabeled videos via normality advantage and self-paced refinement}. In \bibinfo{booktitle}{\emph{CVPR}}. \bibinfo{pages}{13987--13998}.
\newblock


\end{thebibliography}

\clearpage

\appendix
\setcounter{table}{0}
\renewcommand{\thetable}{S\arabic{table}}
\setcounter{figure}{0}
\renewcommand{\thefigure}{SF\arabic{figure}}

\section{Pseudocodes}
As mentioned in the main manuscript, we provide the pseudocodes for the training and testing phases in Algorithm~\ref{algo:pseudo_train} and Algorithm~\ref{algo:pseudo_test}, respectively.
\subsection{Pseudocode for the training phase}
\begin{algorithm}
\caption{CKNN (train)}
  \label{algo:pseudo_train}
  \begin{algorithmic}[1]
  \State \textbf{Input} training videos $\mathbf{D}$, hyperparameters $p$ and $\tau$
  \State \textbf{Output} feature banks $( \mathcal{M}^{app}, \mathcal{M}^{mot} ) $
  \State $\mathcal{B}$ $\leftarrow$ $\left \{ \right \}$
  \For{$v_i$ in $\mathbf{D}$}   \Comment{Object detection}
  \ \For {$f_t$ in $v_i$}
  \ \ \ \ \State $\mathcal{B} \leftarrow \mathcal{B} \cup \texttt{ObjDetect}\left( f_{t} \right)$ \Comment{Eq.~\textred{3}}
  \EndFor
  \EndFor
  \State $\hat{\mathcal{S}}^{app} \leftarrow \left \{ \right \}$, $\hat{\mathcal{S}}^{mot}$ $\leftarrow$ $\left \{ \right \}$     \Comment{Pseudo-anomaly scores}
  \For{$b_{(t,m)}$ in $\mathcal{B}$}
  \ \ \State $\hat{s}^{app}_{(t, m)} \leftarrow f^{app}(b_{(t, m)})$ \Comment{Eq.~\textred{4}}
  \ \ \State $\hat{s}^{mot}_{(t, m)} \leftarrow f^{mot}(b_{(t, m)})$
  \ \ \State $\hat{\mathcal{S}}^{app} \leftarrow \hat{\mathcal{S}}^{app} \cup \{ \hat{s}^{app}_{(t, m)}  \}$  
  \ \ \State $\hat{\mathcal{S}}^{mot} \leftarrow \hat{\mathcal{S}}^{mot} \cup \{ \hat{s}^{mot}_{(t, m)}  \}$
  \EndFor
  \State $\mathcal{D}^{app} \leftarrow \left \{ \right \}$, $\mathcal{D}^{mot} \leftarrow \left \{ \right \}$     \Comment{Feature banks}
  \For{ $b_{(t,m)}$ in $\mathcal{B}$ }
  \ \ \If{ $\hat{s}^{app}_{(t, m)} < \texttt{percentile}(\hat{\mathcal{S}}^{app}, \tau)$ }
  \ \ \ \ \State  $\mathcal{D}^{app} \leftarrow \mathcal{D}^{app} \cup \left \{ \phi^{app}(b_{(t,m)}) \right \}$  \Comment{Eq.~\textred{5}}
  \ \ \EndIf
  \ \ \If{ $\hat{s}^{mot}_{(t, m)} < \texttt{percentile}(\hat{\mathcal{S}}^{mot}, \tau)$ }
  \ \ \ \ \State  $\mathcal{D}^{mot} \leftarrow \mathcal{D}^{mot} \cup \left \{ \phi^{mot}(b_{(t,m)}) \right \}$  \Comment{Eq.~\textred{5}}
  \ \ \EndIf
  \EndFor  
  \State $\mathcal{M}^{app} \leftarrow \texttt{rand}(\mathcal{D}^{app}, p)$ \Comment{Feature bank compression}
  \State $\mathcal{M}^{mot} \leftarrow \texttt{rand}(\mathcal{D}^{mot}, p)$ \Comment{Eq.~\textred{6}}
  \State \textbf{return} $ (\mathcal{M}^{app}, \mathcal{M}^{mot}) $
  \end{algorithmic}
\end{algorithm}

\textbf{Object detection}
The \texttt{for} loop in the lines 4-8 of Algorithm~\ref{algo:pseudo_train} performs the object detection.

\textbf{Object cleansing}
The object cleansing process is executed in lines 9-24.
In this step, pseudo-anomaly scores are estimated for each object in lines 9-15.
Using the two types of pseudo-anomaly scores, the corresponding object sets are cleansed in lines 16-24.

\textbf{Feature bank compression}
Optionally, we randomly sample $p$\% of features in the cleansed object sets, as described in lines 25-26.
The resulting features banks, $\mathcal{M}^{app}$ and $\mathcal{M}^{mot}$, are the outputs of the training procedure.

\subsection{Pseudocode for the inference phase}

\begin{algorithm}[H]
\caption{CKNN (infer)}
  \label{algo:pseudo_test}
  \begin{algorithmic}[1]
  \State \textbf{Input} testing video $v$, feature banks $(\mathcal{M}^{app}, \mathcal{M}^{mot})$, and hyperparameter $k$
  \State \textbf{Output} list of anomaly scores $a$
  \State $a$ $\leftarrow$ \texttt{zeros}$\left( v.\mathrm{length}\right)$  \Comment{list of anomaly scores}
  \For {$f_t$ in $v$}    
  \ \State $\mathcal{B}_t \leftarrow \texttt{ObjDetect}\left(f_t\right)$  \Comment{Object detection}
  \For {$b_{(t, m)}$ in $\mathcal{B}_t$}  
  \ \ \State $s^{app}_{(t, m)} \leftarrow \mathrm{kNN}(\phi^{app}(b_{(t, m)}), \mathcal{M}^{app}, k)$  \Comment{Eq.~\textred{7}} 
  \ \ \State $s^{mot}_{(t, m)} \leftarrow \mathrm{kNN}(\phi^{mot}(b_{(t, m)}), \mathcal{M}^{mot}, k)$
  \ \ \State $s_{(t, m)} \leftarrow \texttt{normal}(s^{app}_{(t, m)}) + \texttt{normal}(s^{mot}_{(t, m)})$  
  \EndFor
  \State $a$[$t$] $\leftarrow \underset{m}{\mathrm{max}} ~ {s_{(t, m)}}$
  \EndFor  
  \State $a$ $\leftarrow$ $\texttt{smooth}(a)$
  \State \textbf{return} $a$
  \end{algorithmic}
\end{algorithm}

The inference phase of CKNN begins with the object detection (line 4-5 of Algorithm~\ref{algo:pseudo_test}).
Following feature extractions and k-nearest neighbor algorithm (lines 7-8) estimate anomaly scores for each object.
The maximum object anomaly score in each frame becomes the frame-wise anomaly score (line 11), and Gaussian smoothing in line 13 is applied to ensure temporal smoothness.

\clearpage
\onecolumn
\section{Additional results}
\subsection{More CKNN Outputs}
\begin{figure*}[h]
    \centering    
    \includegraphics[width=\textwidth]{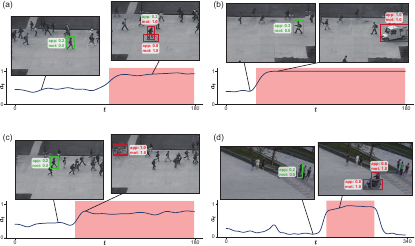}
    \vspace{10pt}
    \caption{\textbf{CKNN detects abnormal objects in test videos from (a-c) Ped2 and (d) STC datasets.} At the bottom of each figure, the estimated anomaly scores for each frame are plotted, with red shades indicating frames labeled as anomalies in the ground truth annotations. Note that only the most distinctive bounding boxes are displayed for a more concise visualization. The four examples demonstrate successful detection of abnormal objects using CKNN. High appearance anomaly scores for the objects show the model's ability to distinguish anomalies based on their appearance. The bicycles in (a, c), car in (b), and motorcycle in (d) receive high appearance anomaly scores because they look abnormal. Their fast movement also results in high motion anomaly scores. The appearance anomaly score for the person riding the bicycle in (a) is low because the bounding box only contains the person and not the bicycle, resulting in a normal appearance.}
    \vspace{-10pt}
    \label{fig:more_examples_obj}
\end{figure*}

\begin{figure*}[h]
    \centering    
    \includegraphics[width=\textwidth]{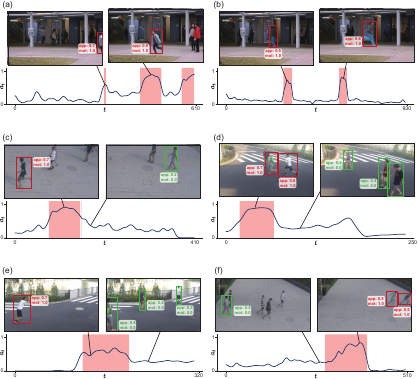}    
    \vspace{10pt}
    \caption{\textbf{CKNN detects abnormal behaviors in test videos from (a-b) Avenue and (c-f) STC datasets.} Examples (a) and (b) demonstrate that CKNN can easily detect abnormal behaviors, such as a jumping child and a running man. The plotted score suddenly spikes and returns to normal as the abnormal behavior is detected and disappears. The STC examples provided in (c-f) present CKNN detecting running or jumping people in frames with multiple objects, showing its successful performance in complex scenarios.}
    \vspace{-10pt}
    \label{fig:more_examples_behavior}
\end{figure*}

In the following figures, we present additional examples of CKNN outputs.
Fig.~\ref{fig:more_examples_obj} shows the results in scenarios containing abnormal \textit{objects}, such as a car, while Fig.~\ref{fig:more_examples_behavior} displays results in scenarios with abnormal \textit{behaviors}, such as a running person.

\vspace{35pt}

\clearpage

\twocolumn
\clearpage

\subsection{Ablation Study}
\subsubsection{Ablation Study1: Employing different modules}
\begin{figure}[H]
\centering
\setlength\tabcolsep{2pt}    
{\renewcommand{\arraystretch}{1.33}%
\captionof{table}{Ablation study.}
\label{table:ablation2}
\begin{tabular}{c|c c|c c} 
\hline %
 & Ped2 & Ped2 & AVE & AVE \\ %
 & P & M & P & M \\
 \hline \hline %
SOTA & 95.7 & 98.1 & 88.7 & 93.0 \\  \hline %
$\phi^{app}=$ ResNet & 99.6 & 99.5 & 89.1 & 93.5 \\ \hline
$f^{app}=$ kNN & 99.2 & 98.6 & 88.9 & 93.1 \\
$f^{app}=$ GMM & 99.4 & 98.6 & 87.1 & 93.9 \\ \hline
$f^{mot}=$ kNN & 99.9 & 99.7 & 92.4 & 93.5 \\
$f^{mot}=$ AE & 96.5 & 97.3 & 93.2 & 92.8 \\ \hline
\textbf{CKNN (Ours)} & \textbf{99.8} & \textbf{99.7} & \textbf{94.0} & \textbf{94.1} \\  \hline
\end{tabular}
}
\end{figure}

In Table 6 of the manuscript, we presented the anomaly detection performances of CKNN on STC dataset with variations in each component.
In Table~\ref{table:ablation2}, we present the results of a similar ablation study on Ped2 and AVE datasets.
As seen in Table 4 of the manuscript, CKNN consistently delivers high detection performance when individual components are altered.

\begin{figure}[h]
    \centering    
    \includegraphics[width=0.5\linewidth]{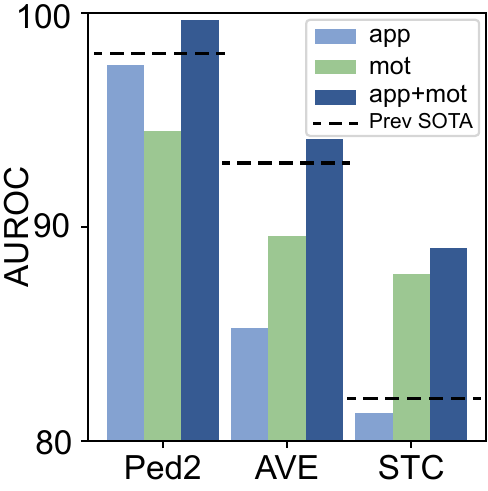}
    \vspace{10pt}
    \caption{The contribution of appearance and motion.}    
    \vspace{-10pt}
    \label{fig:appmot}
\end{figure}

\subsubsection{Ablation Study2: Appearance and motion}
In Fig. 12 of the main manuscript, we illustrated the anomaly detection performance of CKNN when using appearance and motion anomaly scores separately in partial mode.
We also present results from a similar ablation study employing the merge mode evaluation scenario in Fig.~\ref{fig:appmot}.
As observed in Fig. 12, the appearance and motion display a complementary synergy, and utilizing both scores enhances the anomaly detection performance.

\newpage
\section{Experimental Setup}

\textbf{Implementation details}
The hand-crafted optical flow feature extractor~\cite{attr2022arxiv}, denoted as $\phi^{mot}$ in the main manuscript, outputs 8-dimensional vectors in the Avenue, STC, and UBn datasets, while a scalar value is used Ped2 dataset, following the original approach.

As described in the main manuscript, estimating the appearance pseudo-anomaly score requires an autoencoder (AE).
We designed the input and output of the AE to be gray-scale, so we applied RGB-to-Gray conversion as a preprocessing step.
The encoder part of the AE comprises six convolutional layers and four max-pooling layers, with the number of channels in the convolutional layers starting from 32 and increasing to 128 for deeper layers.
The decoder part of the AE has a mirrored shape of the encoder and uses transposed convolutional layers.
We used the leaky ReLU activation function~\cite{leakyrelu} with a leakage parameter value of 0.1 and trained the AE for ten epochs using the Adam~\cite{adam} optimizer with a learning rate of 0.001 and a batch size of 64.
For more detailed information on the architecture of the AE, please refer to the provided code, which is available online\footnote{\url{https://github.com/nuclearboy95/Anomaly-Detection-CKNN-PyTorch}}.
Instructions for preparing the dataset and required libraries are described in \texttt{README.md} file.

In unsupervised video anomaly detection scenarios, the test split of a dataset is used in both training and testing phases.
Therefore, during the testing phase, the nearest neighbor (NN) search of the k-NN algorithm may result in the query data itself being identified as the nearest neighbor.
To avoid this issue when computing the anomaly score in Eq.~\textred{1}, we excluded one exact match (if it exists) from each NN search result.

\textbf{Libraries}
For the implementation, we used pytorch~\cite{pytorch} and scikit-learn~\cite{scikitlearn} libraries for the training of the AE and Gaussian Mixture Model, respectively.
k-NN search was implemented using faiss~\cite{faiss} library.










\balance

\end{document}